**EEG-based AI-BCI Wheelchair Advancement: Transformer-Based Learning with Motor Imagery for Brain Computer Interface**

**Bipul Thapa[a], Biplov Paneru[b], Bishwash Paneru[c], Khem Narayan Poudyal[c]**

[a]Department of Computer Science and Engineering, Kathmandu University, Kavre, Nepal

[b]Department of Electronics and Communication Engineering, Nepal Engineering College, Pokhara University, Bhaktapur, Nepal

[c] Department of Applied Sciences and Chemical Engineering, Institute of Engineering, Pulchowk Campus, Tribhuvan University, Lalitpur, Nepal

***Corresponding author 1**: Bipul Thapa[a]

***Corresponding author 2**: Biplov Paneru[b]

Email: bipulthapa23@gmail.com, thebplvstar001@gmail.com

***Abstract***

This paper presents an Artificial Intelligence (AI) integrated approach to Brain-Computer Interface (BCI)-based wheelchair development, utilizing a motor imagery right-left-hand movement mechanism for control. The system is designed to simulate wheelchair navigation based on motor imagery right and left-hand movements using electroencephalogram (EEG) data. A pre-filtered dataset, obtained from an open-source EEG repository, was segmented into arrays of 19x200 to capture the onset of hand movements. The data was acquired at a sampling frequency of 200Hz. The system integrates a Tkinter-based interface for simulating wheelchair movements, offering users a functional and intuitive control system. We propose TFormerEEG, a Transformer-driven deep learning architecture, for motor imagery EEG classification. The model achieves a test accuracy of 93.04% compared with various machine learning baseline models, including XGBoost, EEGNet, and an EEG-Deformer model. The TFormerEEG achieved a mean accuracy of 91.18% through stratified cross-validation, showcasing the effectiveness of this model.

***Keywords***: *Brain Computer Interface (BCI), Motor Imagery, Transformer-based, Raspberry Pi, Electroencephalogram (EEG), TformerEEG*

## 1. Introduction

Brain-Computer Interfaces (BCIs) are advanced systems that establish direct communication between the human brain and external devices. In recent years, BCIs have attracted significant research attention due to their potential to assist individuals with mobility impairments, providing novel pathways for restoring autonomy. Among the many assistive applications being explored, BCI-controlled wheelchairs represent a promising solution for enabling mobility in individuals with severe motor disabilities [1, 2].

In such systems, effective operation depends on the accurate interpretation of neural signals, which requires reliable analysis of brain activity using state-of-the-art neurotechnology. These neural activity patterns are captured through multiple advanced techniques that measure and interpret signals generated by the brain [3]. One important category of signals used in these systems is biopotentials, which are generated through electrochemical processes in the human body and have been extensively studied for their applicability in human-machine interfaces (HMIs), including motorized wheelchair systems. These signals enable direct interaction between neural activity and external devices.

Electroencephalography (EEG), a non-invasive modality that captures cortical brain activity, is widely employed to develop such systems [4]. With growing research interest in EEG-based BCIs and brain disorder diagnostics, EEG technologies are expected to have a significant societal impact. However, existing systems remain insufficient for fully automated detection and integration within BCI-enabled frameworks [5]. Prior experiments involving wheelchairs operated via brain signals have demonstrated the feasibility of translating neural intent into motor commands, confirming the potential of EEG-based control methods in assistive mobility systems [6].

The COVID-19 pandemic further underscored the urgent need for innovative assistive technologies to support individuals with severe disabilities in their daily lives. BCIs utilizing EEG signals facilitate greater independence for individuals facing serious health challenges, improving their well-being, while also enabling users to operate various devices without requiring physical limb movement, enhancing the quality of life of elderly individuals and those with physical impairments [7, 8].

In response to these challenges, intelligent wheelchair technologies have been increasingly explored to improve mobility for individuals with physical impairments. The application of BCI techniques in the development of electric wheelchairs has attracted considerable interest due to their adaptability for individuals with physical disabilities [9]. However, most current BCI systems are designed to control a single assistive device, such as a robotic arm, prosthetic limb, or wheelchair. Moreover, some systems have overlooked the specific needs of users with severe physical limitations, particularly those lacking sufficient muscular strength or experiencing paresis, who may be unable to maneuver conventional wheelchairs [10]. This limitation highlights the need for alternative control mechanisms that do not rely on physical limb movement, such as motor imagery-based interaction. In practice, real-world daily activities often require coordination of multiple assistive capabilities, which can only be achieved by integrating several robotic systems into a unified framework [11].

Moreover, these systems remain in the research stage and are often constrained by key limitations, including inadequate attention to users' mental activity, variability in neural behavior across different environments, and suboptimal accuracy in classification tasks [12]. Furthermore, the convergence of BCI and smart wheelchair (SW) technologies introduces additional concerns related to system security, privacy, and operational safety, which remain insufficiently addressed in existing large-scale studies [13]. At the same time, prior research has demonstrated the feasibility of using non-invasive BCIs to control both real-world devices (e.g., wheelchairs, quadcopters) and virtual objects (e.g., computer cursors, virtual helicopters) [14].

Despite these advancements, current EEG-based BCI systems face critical limitations in practical deployment. Many systems suffer from inadequate adaptability to user-specific neural dynamics and limited robustness under real-world environmental variability. Additionally, classification models often struggle to achieve the level of accuracy required for precise and reliable control. The lack of integrated support for users with extreme physical impairments—such as those unable to generate muscle responses—combined with the focus on single-device control, restricts the real-world utility of these systems. These persistent issues point to a significant research gap in developing BCI wheelchair systems that are simultaneously accurate, adaptable, and user-centric.

Recent advancements in artificial intelligence (AI) have significantly enhanced the capability of interpreting EEG signals for automated control systems. The integration of AI into signal processing is revolutionizing robotics and automation by minimizing the need for direct human intervention. As a result, the ability to decode EEG signals effectively has opened new opportunities for automating machinery through BCI frameworks [15].

Motivated by these challenges, this paper proposes a BCI-based wheelchair control framework driven by EEG signals associated with motor imagery. In particular, we propose TFormerEEG, a transformer-based model, to enhance EEG signal classification performance for motor imagery-based BCI wheelchair control. The proposed approach is evaluated against several baseline models, including XGBoost, EEGNet, and EEG-Deformer models, using tailored hyperparameter optimization techniques.

The main contribution of our work is summarized as follows:

1. Development of a motor imagery EEG-based BCI framework for wheelchair control.
2. Proposed TFormerEEG, a Transformer-based deep learning architecture for improved motor imagery EEG classification.
3. Comparative evaluation of the proposed model against several baseline approaches, including XGBoost, EEGNet, and EEG-Deformer models.

The remainder of this paper is organized as follows. Section 2 reviews the existing literature on EEG-based BCI systems and related assistive wheelchair technologies. Section 3 presents the proposed methodology, including the dataset description, preprocessing procedures, and the TFormerEEG architecture. Section 4 presents the experimental results and comparative evaluation of the proposed approach with baseline models, along with a discussion of the study limitations and potential future research directions. Finally, Section 5 concludes the paper.

## 2. Literature Review

Several studies have examined the evolution of the BCI systems for assistive mobility [7, 9]. In this section, we discuss some existing related works.

### 2.1 Commercial EEG Headset-Based Systems

Several studies have explored the development of BCI-based systems using commercially available EEG headsets due to their affordability and ease of deployment. Swee et al. [3] developed a brainwave-controlled wheelchair using the Emotiv EPOC headset, where EEG signals are processed on a personal computer to generate control commands transmitted wirelessly to the wheelchair. Similarly, Maksud et al. [8] developed a low-cost EEG-based electric wheelchair using the Neurosky MindWave Mobile device, which detects attention levels and eye blinks to enable control without physical input. Dev et al. [16] designed an EEG-based brain-controlled wheelchair utilizing the NeuroSky MindWave headset, where movement is governed by the user's attention level and activation is triggered via double eye blinks.

### 2.2 Artifact-Based Control Paradigms

Many BCI wheelchair systems rely on artifact-based signals such as eye blinks, facial expressions, or muscle contractions. Rao et al. [17] proposed a smart wheelchair system that utilizes EEG signals captured via a Brainsense headset, leveraging artifact-based inputs such as eye blinks for directional control. Similarly, Tippannavar et al. [18] proposed a BCI-enabled wheelchair system utilizing EEG signals alongside facial movements, eye blinks, and muscle contractions, with Raspberry Pi selected for its wireless connectivity, remote update capability, and parallel processing support. Awais et al. [19] developed a brain-controlled wheelchair system using the NeuroSky MindWave headset, integrating joystick and Android-based remote-control interfaces to provide multiple control modes.

### 2.3 Motor Imagery-Based BCI Systems

Motor imagery (MI) represents one of the most widely studied paradigms in BCI research due to its ability to capture neural representations of voluntary movement without requiring physical execution. Tanaka et al. [20] investigated EEG-based directional control of an electric wheelchair using a recursive training algorithm to generate recognition patterns from brain signals. Experimental results confirmed the feasibility of controlling wheelchair movement solely through EEG input. Ghasemi et al. [13] introduced a non-invasive BCI system that interprets user-trained EEG signals to decode specific mental commands for precise wheelchair navigation. The system translates cognitive intentions into directional control, aiming to enhance autonomy for individuals with physical disabilities.

### 2.4 Hybrid Signal BCI systems

Hybrid BCI systems integrate multiple physiological signals to enhance classification accuracy and system reliability. Welihinda et al. [10] proposed a cost-effective manual-to-powered wheelchair conversion kit using a hybrid EEG-EMG control system trained with an LSTM network. The system achieved 97.3% overall accuracy, enabling ergonomic and biopotential-based control tailored for elderly and disabled users.

Huang et al. [11] introduced a hybrid BCI system combining EEG and EOG signals to control both a wheelchair and an integrated robotic arm, enabling complex task execution through motor imagery and ocular movements. Their system demonstrated high control accuracy, with successful real-world application in a mobile self-drinking experiment. Zhao et al. [21] proposed a hybrid EEG-EMG BCI framework for motor imagery classification using an ensemble learning model to enhance signal classification accuracy.

### 2.5 Alternative EEG control paradigms

Several studies have explored alternative EEG paradigms for BCI-based wheelchair control. Ming et al. [22] developed a system based on alpha-wave blocking triggered by eye closure, achieving a success rate of 93.7% and an information transfer rate of 12.54 bits/min. Singla and Haseena [23] proposed an SSVEP-based wheelchair control system using flickering visual stimuli and FFT-based EEG analysis, where a One-Against-ALL SVM classifier achieved superior performance compared to ANN models.

Other works have focused on system-level architectures and hardware platforms. Liu et al. [24] introduced a brain-controlled wheelchair integrating computer vision and augmented reality for target-based navigation, achieving over 83% accuracy. Hassin and Khan [25] developed NeuroSpy, a portable IoT-enabled biomedical recorder capable of capturing EEG, ECG, and body temperature signals in real time, supporting scalable BCI healthcare applications.

## 2.6 AI-Based Classification Approaches

Recent advances in artificial intelligence have significantly improved EEG signal decoding. Hasan et al. [12] developed a brainwave-controlled wheelchair system using EEG responses to color stimuli, mapping four directions to specific colors and rhythms. Their analysis identified the beta rhythm as most effective, with an ANN classifier achieving the highest accuracy of 82.5%. Qidwai et al. [5] proposed a lightweight EEG signal classification method using fuzzy logic and statistical features, suitable for microcontroller-based brain-controlled wheelchair systems. Their approach targets real-time control in ubiquitous assistive settings, addressing the constraints of compact, memory-limited hardware.

Abiyev et al. [6] designed a brain-actuated wheelchair control system using a fuzzy neural network (FNN) algorithm to classify EEG signals based on user mental activity. The system demonstrated improved directional and speed control accuracy while minimizing misclassification in real-world conditions. Arshad et al. [15] investigated a non-invasive BCI system for robotic arm control using various AI classifiers, finding Random Forest achieved the highest accuracy at 76%, followed by Gradient Boosting at 74%. The study also highlighted individual differences in EEG signal characteristics that influence classification performance.

Zhang et al. [26] explored AI-driven classification methods within a non-invasive BCI system for robotic arm control, identifying Random Forest as the most accurate (76%) among six evaluated models. The study also emphasized the impact of individual EEG signal variability on classification performance, informing intelligent wheelchair interface design for enhanced user adaptability. An et al. [27] introduced a hybrid BCI control system using an LSTM-CNN for EEG feature extraction and an actor-critic model to mitigate noise and cognitive variability. Implemented with the Unicorn Hybrid Black device, the system achieved 93.12% online accuracy and a 67.07 bits/min information transfer rate, outperforming existing BCI control systems.

Paneru et al. [28] developed an EEG-based Brain-Machine Interface for keystroke classification, using ERP segmentation and deep learning architecture, the BiGRU-Attention model. Also, a real-time GUI was implemented to simulate keystrokes for individuals with motor disabilities. Paneru et al. [29] proposed a Stacking Ensemble Technique model for EEG-based robotic car control using motor action signals, demonstrating the effectiveness of ensemble learning for improving classification performance in BCI systems.

## 2.7 BCI Applications in Assistive Robotics

In addition to wheelchair navigation, BCI technologies have been applied to other assistive robotic systems. Meng et al. [14] demonstrated that non-invasive EEG signals could be used to control a multi-degree-of-freedom robotic arm for reach-and-grasp tasks. Their study involved thirteen participants who successfully maintained accurate robotic control over several months of training, demonstrating the long-term viability of EEG-based assistive robotics.

Similarly, Bousseta et al. [4] developed a robotic arm control system using imagined motor tasks captured via an Emotiv EPOC headset. Their approach employed principal component analysis (PCA) and fast Fourier transform (FFT) for feature extraction and a support vector machine classifier, achieving an average accuracy of 85.45%. Shedeed et al. [30] proposed another robotic arm control system based on EEG signals associated with motor tasks, where features extracted using wavelet transforms and multilayer perceptron networks achieved classification accuracy of up to 91.1%.

Maassarani et al. [31] further developed an EEG-based robotic arm control system that interprets mental commands and facial expressions through Cortex and Python-based processing pipelines. These studies demonstrate the broader applicability of EEG-based BCIs in assistive robotics and reinforce the feasibility of translating neural signals into control commands for external devices. These studies demonstrate the feasibility of EEG-based BCIs for controlling assistive robotic systems and highlight the potential of translating neural activity into external device control.

Despite the significant progress in EEG-based BCI systems, several challenges remain that limit their practical deployment in assistive mobility applications. Many existing approaches rely on artifact-driven control mechanisms that utilize non-neural signals, which may lead to unintended command activation and reduce the reliability of the system. Other systems employ hybrid signal BCI frameworks that integrate multiple biosignals to improve classification performance; however, these approaches require additional sensing hardware, increasing system complexity, cost, user setup requirements, and calibration overhead. Furthermore, a large portion of the literature still relies on conventional machine learning techniques or shallow neural architectures that are insufficient for capturing the complex spatiotemporal characteristics of motor imagery EEG signals. These limitations highlight the need for advanced deep learning frameworks capable of learning robust feature representations from EEG data while maintaining a practical single-modality BCI framework for reliable wheelchair control.

The objective of this study is to develop a system that allows users to control wheelchair movements using their brain signals, specifically the movements of their right and left hands. Decoding EEG data based on the right and left-hand motor imagery of users is the first approach to such an application, and by translating it into control commands, the system offers a promising solution for hands-free wheelchair navigation.

## 3. Methodology

### 3.1 Dataset and Preprocessing

#### 3.1.1 Dataset Description

Our work utilizes a publicly available EEG dataset described in [32], designed to explore the neural correlates of motor imagery using a Classical (CLA) paradigm. EEG signals were recorded using the EEG-1200 acquisition system with electrodes positioned according to the international 10–20 configuration. While the complete recording includes 22 channels, only 19 channels corresponding to standard cortical EEG sites were retained for analysis. Channels A1, A2, and X5 were excluded due to their reference or non-cortical roles, ensuring that only task-relevant neural activity was captured.

The dataset is based on the CLA motor imagery paradigm corresponding to the onset of hand movements, in which participants were first shown action signals denoting one of the mental imagery exercises that needed to be done. This protocol allowed for the study of brain activity prior to physical action, focusing on the differentiation between rest, left-hand, and right-hand motor imagery. There are three states of mental imagery Subject E (3St). Left-Right Hand (LRHand) is the mnemonic for the recording session. The EEG signals reflect the motor planning phase that precedes physical keypresses, capturing essential dynamics for BCI research [32].

Each dataset file, as shown in Table 1, contains continuous EEG data, along with a synchronized marker signal that indicates event onsets. A change in the marker signal from 0 to 1 designates the onset of right-hand imagery, whereas a change from 0 to 2 indicates left-hand imagery. This annotation method facilitates precise ERP alignment and classification training.

**Table 1**. EEG dataset

| Dataset Name | Shape of 'data' | Number of Events |
|---|---|---|
| CLASubjectE1601223StLRHand.mat | (664400, 22) | 635 |
| CLASubjectE1601193StLRHand.mat | (664000, 22) | 634 |
| CLASubjectE1512253StLRHand.mat | (667000, 22) | 635 |

Figure 1 shows the raw EEG time-series data from all 22 recorded electrodes (including references), giving an overview of the spatial dynamics across the scalp. Each subplot represents the continuous EEG trace of a single electrode, revealing the rhythmic activity and baseline noise characteristics inherent in the dataset. This visualization serves as an initial quality inspection to identify signal trends, oscillatory behavior, and potential artifacts.

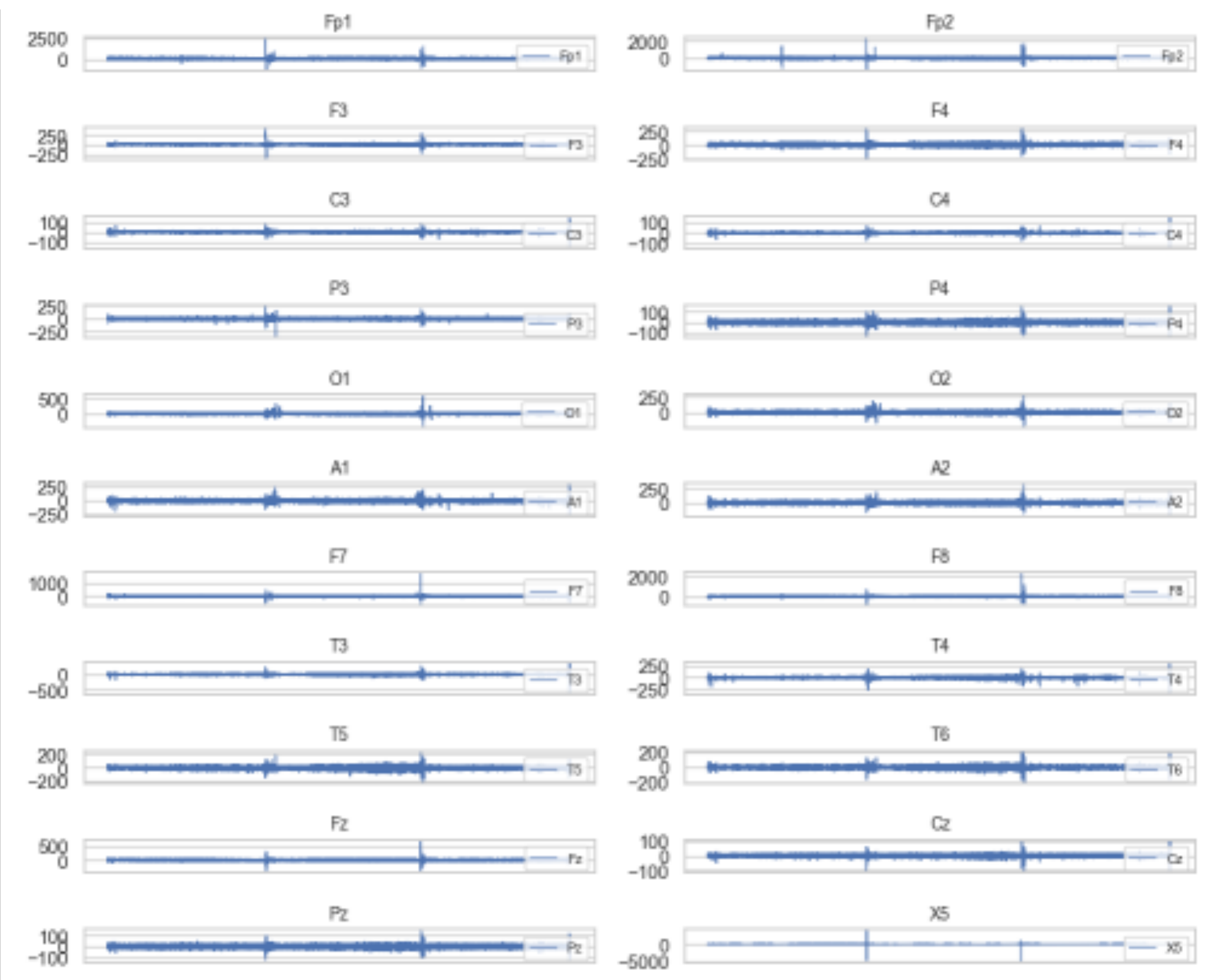

**Fig 1.** Raw channels data plot

### 3.1.2 Preprocessing

Initial preprocessing was performed at the data collection stage that included band-pass filtering between 0.53 Hz and 100 Hz. This range preserves frequencies critical to motor-related EEG activity while attenuating slow drifts and high-frequency noise [32]. In this work, we utilized the recordings as provided to maintain consistency with the published protocol and to evaluate the proposed model under realistic EEG conditions. No additional artifact rejection is applied. Prior to model training, z-score normalization was performed using the StandardScaler method.

Feature extraction is performed on this data window-based segmentation using 19 channels and a sampling frequency of 200, resulting in an array in the form of 19*200. Feature extraction is performed using ERP-based temporal windowing. Around each identified motor imagery onset (right or left hand), a segment of EEG data spanning 2 seconds before to 2 seconds after the event is extracted, yielding a 4-second window across 19 channels. These temporal segments are then flattened into feature vectors for downstream classification tasks. The final dataset contained 3808 samples and 3801 columns. Each sample represents an EEG window extracted around a motor imagery event. The feature space was obtained by flattening EEG signals from 19 channels across a temporal window of 200 samples, resulting in 3800 feature values, with the final column representing the class label.

Each record in the dataset is distinguished by a unique alphanumeric identifier referred to as "id." This identifier serves as a key element for record tracking and management. The "nS" parameter denotes the number of EEG data samples contained within each record, providing insight into the temporal dimension of the recorded neural signals. The "sampFreq" parameter specifies the sampling frequency of the EEG data, representing the rate at which data points are collected per unit of time. The "marker" field encapsulates the eGUI interaction record of the recording session, offering contextual information about user actions during the EEG data acquisition [32].

The final dataset includes three classes corresponding to rest (0), right-hand motor imagery (1), and left-hand motor imagery (2). The class distribution consists of 1904 samples for class 0, 963 samples for class 1, and 941 samples for class 2. The rest class contains a larger number of samples because rest-state EEG segments occur between motor imagery trials during continuous recording. The dataset was divided using stratified sampling into 80% training and 20% testing sets. Additionally, 20% of the training data was used as a validation set during model training.

### 3.1.3 ERP Analysis

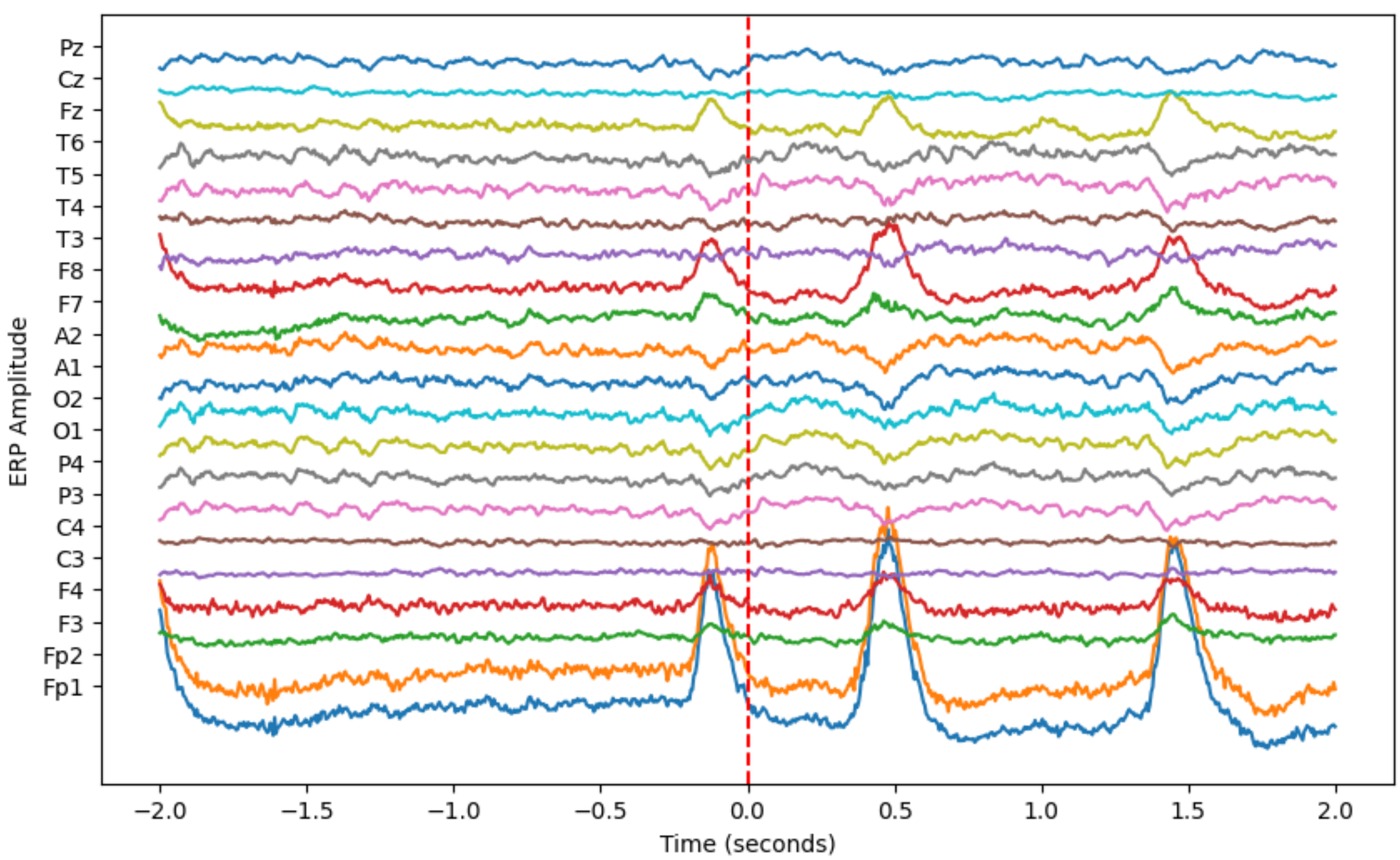

**Fig 2.** Representative event-centered EEG epochs (single-trial)

Figure 2 shows representative EEG epochs aligned to the annotated onset of left-hand motor imagery. The dashed vertical line indicates the event onset (t = 0 s), and the signals illustrate EEG activity across channels within a −2 s to +2 s window.

The ERP provides information about the subject's brain activity during the CLA. The peaks and valleys in the waveforms indicate various electrical activity patterns, and the various colors correspond to various scalp electrodes. For instance, the Pz electrode, which is located at the top of the head, exhibits a positive peak at approximately 300 milliseconds, which is commonly connected to the ERP's P300 component.

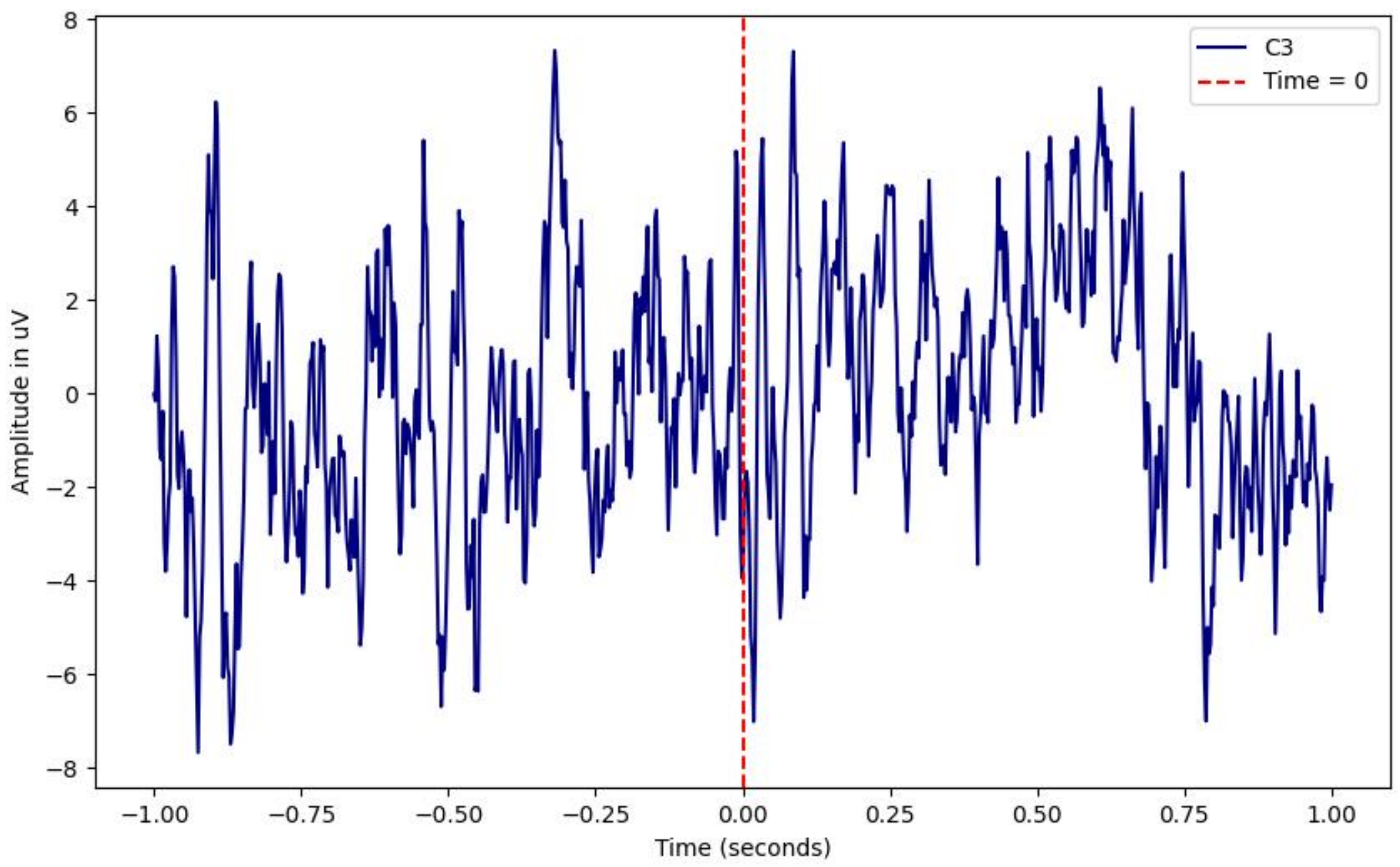

**Fig 3.** ERP plot of channel C3

Figure 3 shows the averaged ERP waveform for channel C3, which is a key site for detecting left-hand motor-related activity. The red dashed line at time zero indicates the onset of motor imagery. The figure captures the temporal evolution of neural responses centered around this onset and provides insight into event-locked cortical activations.

### 3.2 Proposed Model: TFormerEEG

The proposed methodology introduces TFormerEEG, a patch-embedded Transformer architecture that uses a structured token representation and self-attention mechanism to effectively model complex sequential patterns in EEG motor imagery signals, as shown in Figure 4. Specifically, the input EEG feature sequence is divided into fixed-length segments and projected into an embedding space, while the Transformer is employed to capture long-range contextual dependencies among the learned feature representations.

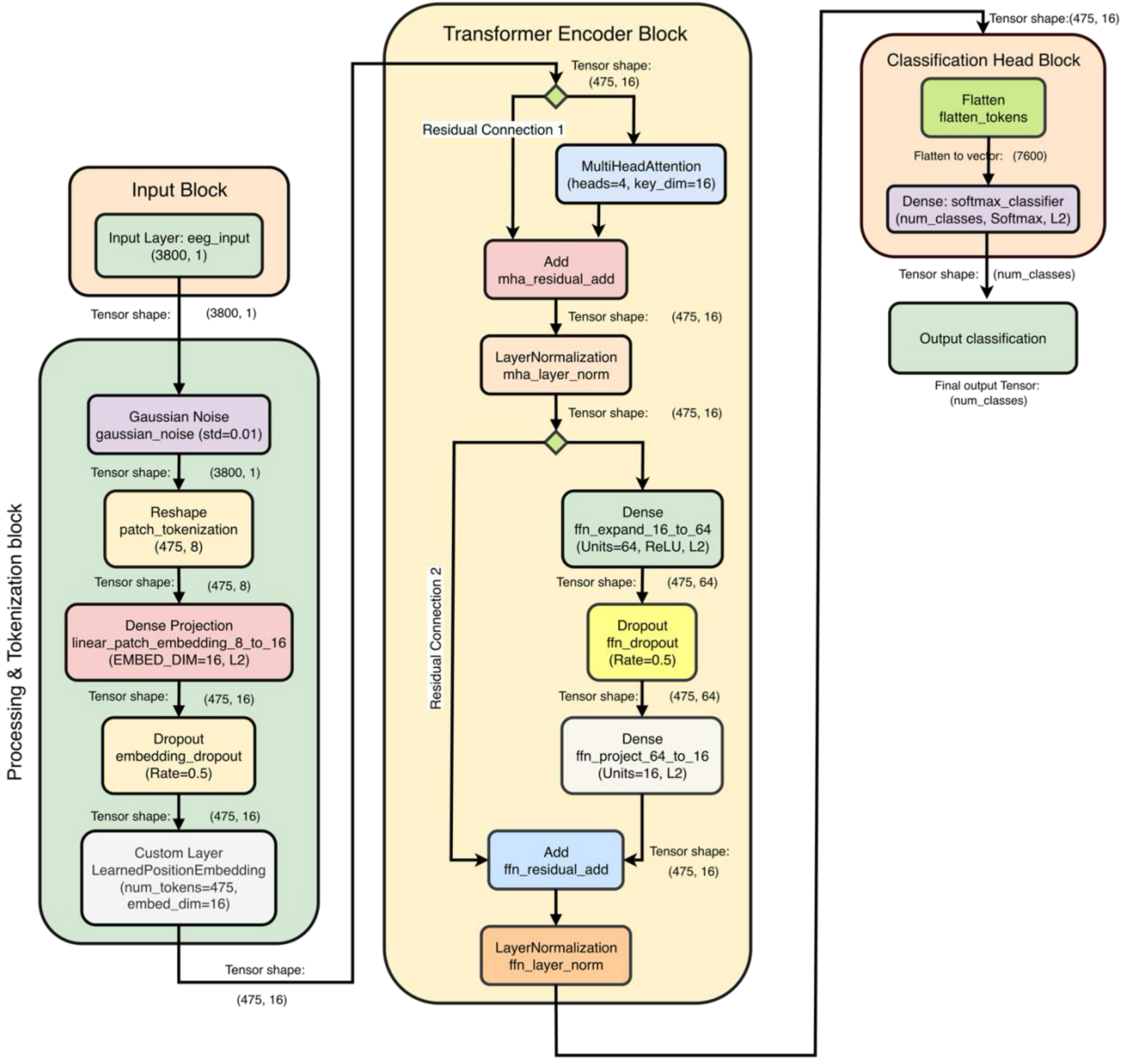

**Fig 4.** Proposed TFormerEEG architecture

Each EEG segment consists of 19 channels sampled over 200 temporal points, forming a 19×200 spatiotemporal representation that is subsequently flattened into a 3800-dimensional input vector. The input is first passed through the TFormerEEG model, where an initial Gaussian noise layer is incorporated to improve generalization and robustness by introducing mild stochastic perturbations during training. The resulting feature sequence is then reshaped into non-overlapping fixed-length segments of size 8, producing a sequence of shape (475,8).

*Transformer Architecture:* Following fixed-length segmentation and embedding, the input to the Transformer encoder has shape (475,16). This sequence is taken as 475 tokens, with each token corresponding to one segment of the flattened EEG feature vector represented in a 16-dimensional embedding space. The only segmentation applied before the Transformer is the division of the 3800-dimensional EEG vector into consecutive segments of length 8. No additional grouping or projection is performed before the embedded sequence is passed to the multi-head self-attention layer.

To preserve the temporal ordering of the EEG sequence, a learned positional encoding is incorporated prior to the Transformer encoder. Specifically, a trainable positional embedding matrix of shape (475, 16) is learned, where each position index is mapped to a 16-dimensional vector. These positional vectors are added elementwise to the token embeddings, ensuring that the model retains information about the sequential structure of the EEG signal despite the permutation-invariant nature of self-attention. This enables the model to effectively capture temporal dependencies critical for motor imagery classification.

The Transformer component consists of a single encoder block. This encoder follows the standard Transformer structure consisting of multi-head self-attention followed by a position-wise feed-forward network with residual connections and layer normalization. Within this block, multi-head self-attention with four attention heads is applied to model global dependencies across the sequence. In the multi-head attention module, the key_dim = 16 specifies the dimensionality of the projected query and key vectors for each attention head. The attention output is combined with the input through a residual connection, followed by layer normalization to ensure stable and efficient training.

The attention-enhanced features are further processed by a position-wise feed-forward network. This sublayer follows a two-step projection: the feature dimension is first expanded from 16 to 64 using a ReLU activation and then projected back to 16. This design introduces additional representational capacity while maintaining compatibility with residual connections. Dropout is applied within the feed-forward network for regularization, followed by a second residual connection and layer normalization.

The proposed TFormerEEG employs a single Transformer encoder block. The rationale for this design is motivated by three considerations. First, the fixed-length segmentation and embedding stage produces a structured feature sequence suitable for attention-based modeling. Second, given the relatively limited size of the EEG dataset, deeper Transformer architectures increase the risk of overfitting without yielding significant performance gains. Third, a lightweight Transformer configuration enables efficient deployment on resource-constrained platforms, such as embedded systems and Raspberry Pi-based implementations, which are relevant for real-time BCI applications.

At the final stage, the refined feature sequence is flattened and passed to a dense SoftMax layer for multi-class classification. By using fixed-length EEG segmentation, token embedding, learned positional encoding, and attention-based global dependency modeling, the proposed TFormerEEG provides an effective and computationally efficient framework for EEG-based motor imagery classification. The detailed layer configuration is presented in Table 2, and hyperparameter settings are reported in Table 6.

**Table 2.** Detailed layer configuration and output shapes of the proposed TFormerEEG

| **Layer** | **Configuration** | **Output Shape** |
|---|---|---|
| Input Layer | EEG feature input | (seq_len, 1) |
| Gaussian Noise | σ = 0.01 | (seq_len, 1) |
| Reshape | num_tokens = seq_len / 8, patch_size = 8 | (num_tokens, 8) |
| Patch Embedding (Dense) | Dense(16), L2=0.2 | (num_tokens, 16) |
| Dropout | rate = 0.5 | (num_tokens, 16) |
| Positional Encoding | Learned Embedding | (num_tokens, 16) |
| Multi-Head Self-Attention | 4 heads, key_dim = 16 | (num_tokens, 16) |

| Residual Add | Skip connection | (num_tokens, 16) |
|---|---|---|
| Layer Normalization | Default configuration | (num_tokens, 16) |
| Feed Forward Dense-1 | Dense(64, ReLU), L2=0.3 | (num_tokens, 64) |
| Dropout | rate = 0.5 | (num_tokens, 64) |
| Feed Forward Dense-2 | Dense(16), L2=0.3 | (num_tokens, 16) |
| Residual Add | Skip connection | (num_tokens, 16) |
| Layer Normalization | Default configuration | (num_tokens, 16) |
| Flatten | Flatten sequence representations | (num_tokens × 16) |
| Output Layer | Dense(num_classes, Softmax), L2=0.2 | (num_classes) |

### 3.3 Baseline

To evaluate the effectiveness of the proposed TFormerEEG model, we compare it with baseline models: XGBoost, EEGNet [33], and an EEG-Deformer model [34]. XGBoost serves as a strong classical machine learning benchmark widely used for structured feature-based classification tasks. EEGNet is a compact convolutional neural network specifically designed for EEG signal analysis and has been widely adopted as a benchmark in EEG-based studies. In contrast, EEG-Deformer combines self-attention mechanisms to capture global dependencies with fine-grained convolutional branches to preserve local temporal details, processing in a hierarchical coarse-to-fine manner across multiple layers [34].

All baseline models were trained and evaluated using the same preprocessing pipeline, training approach, and evaluation metrics as the proposed method to ensure a fair comparison. This comparison enables a systematic evaluation of the proposed architecture against advanced architectures in EEG classification tasks.

### 3.4 Cross-Validation and Accuracy

A 10-fold stratified cross-validation technique is applied to evaluate each model's performance. This method ensures that the models are trained and tested on different subsets of the data, reducing the likelihood of overfitting and providing a more reliable estimate of the model's accuracy.

***User Interface Design: Wheelchair Simulation Interface***

A Tkinter-based graphical user interface (GUI) is developed to simulate the control of a wheelchair using the classified EEG signals. Upon receiving EEG signals, the system classifies the signals as either right-hand or left-hand movements, which are then translated into corresponding movements of the wheelchair (e.g., moving left or right). The interface allows users to interact with the system in real-time, providing a visual representation of wheelchair movements based on the decoded right and left-hand commands.

***Real-Time Simulation***

The system enables real-time processing of EEG signals and updates the wheelchair's movements accordingly. This offers a hands-free control mechanism for wheelchair-bound individuals, with the potential for real-world applications in assistive technology. The overall proposed workflow is presented in Figure 5.

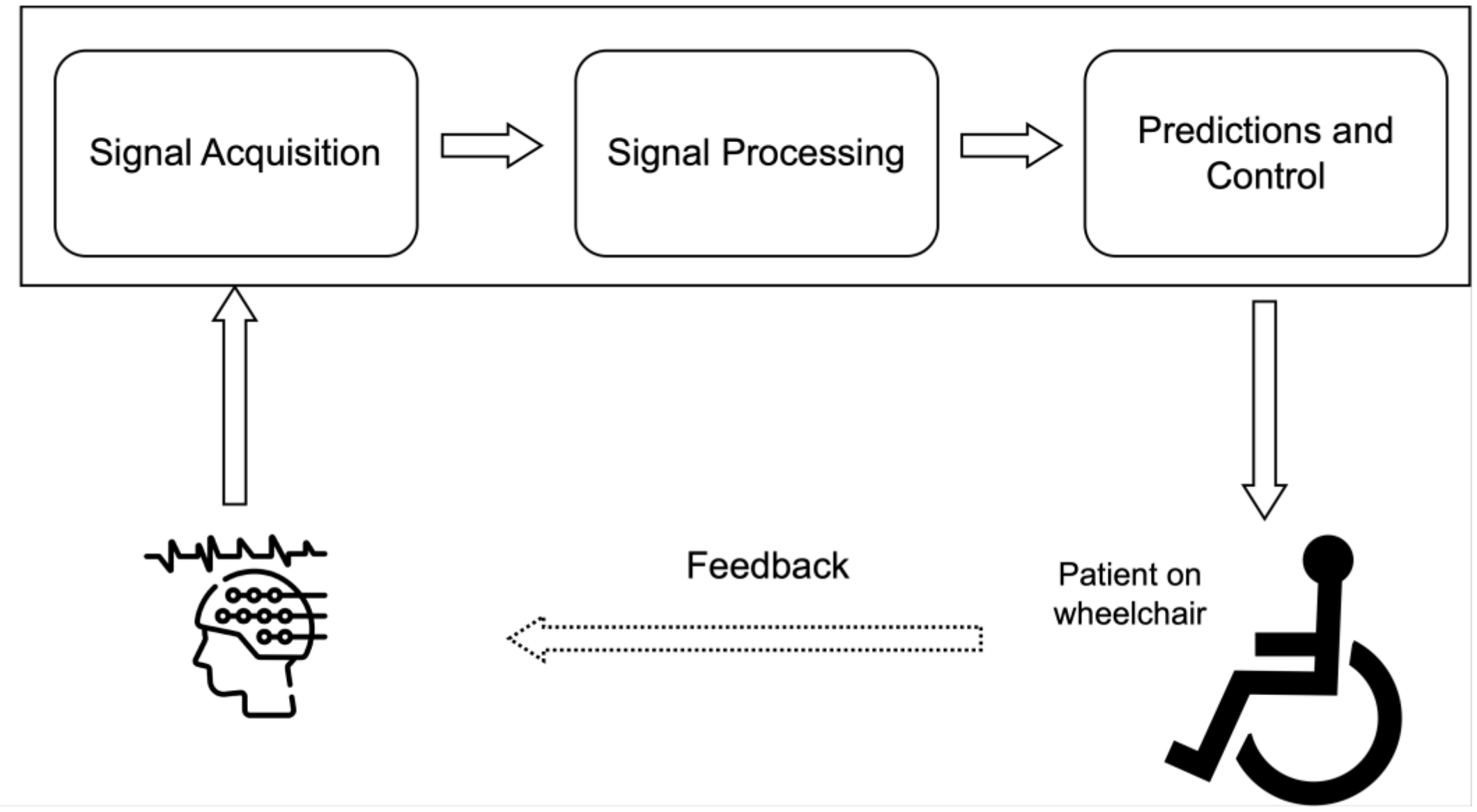

**Fig 5.** Proposed workflow

## 3.5 Proposed circuit

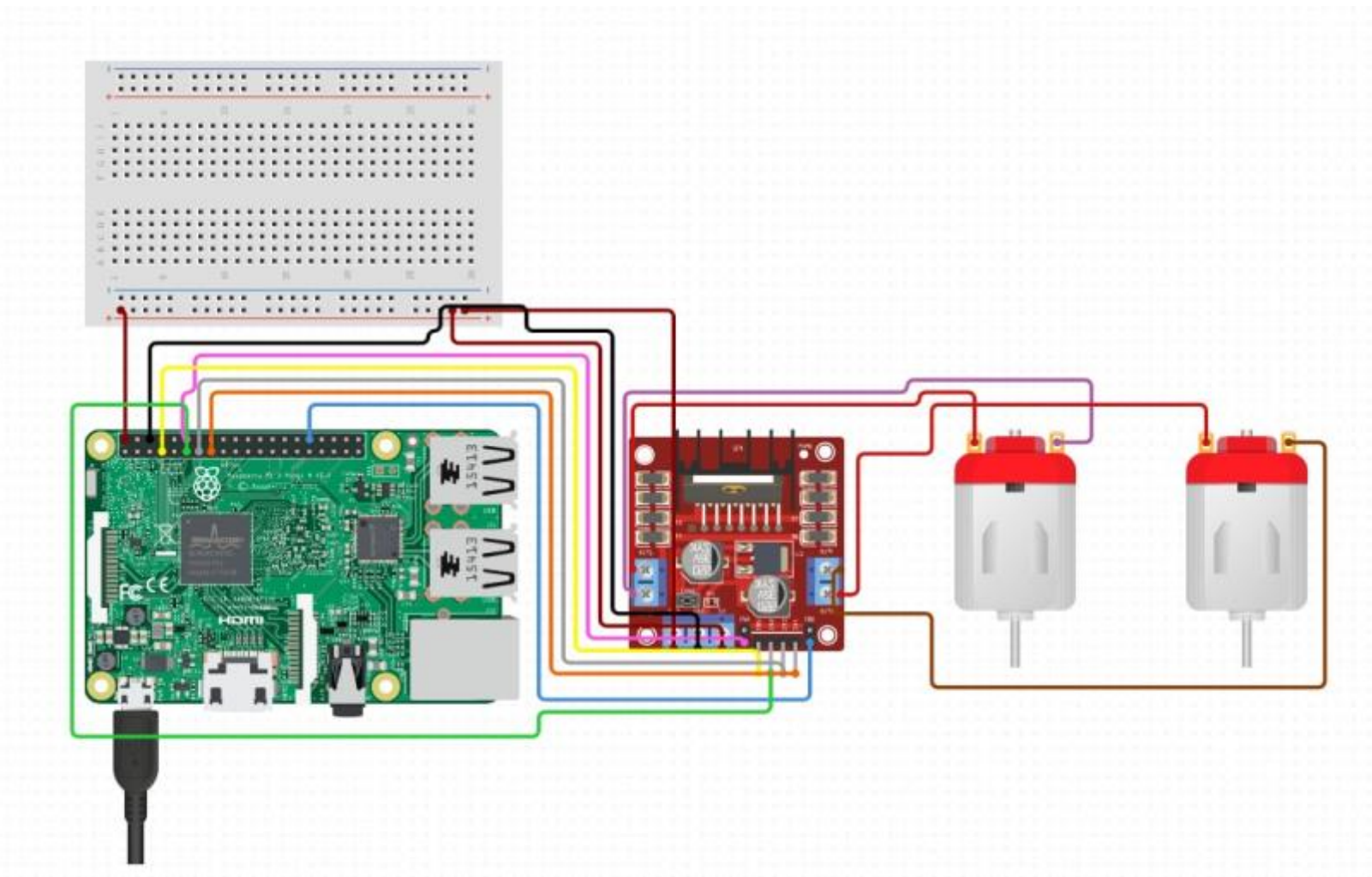

**Fig 6.** Proposed system circuit

The proposed system circuit illustration is shown in Figure 6, the motor driver can be controlled by the model whenever the system detects motor imagery hand movement with the help of an EEG headset utilized by a patient. With this approach, motors can be operated to move the wheelchair with the help of the motor imagery hand movements.

### 3.6 Evaluation Metrics:

**i. Accuracy:**

The percentage of true positive and true negative predictions made out of all the forecasts, which shows how accurate the model is overall.

 $Accuracy = \frac{TP+TN}{(FP+FN+TP+TP)}$.................. Eq. 1

**ii. Precision:**

The accuracy of the positive class predictions is indicated by the ratio of true positive predictions to all predicted positives.

 $Precision = \frac{TP}{(FP+TP)}$ .......................... Eq. 2

**iii. Recall:**

It is a measure of how well the model detects positive cases, expressed as the ratio of true positive predictions to real positives in the dataset.

 $Recall = \frac{TP}{(FN+TP)}$.......................... Eq. 3

**iv. F1-score:**

In the cases, where class distributions are uneven, the harmonic mean of accuracy and recall offers a balanced assessment of a model's performance on positive classes.

$$F1\ Score = 2 \times \frac{Precision * Recall}{Precision + Recall} \quad \text{Eq. 4}$$

**v. AUC-ROC (Area Under the Receiver Operating Characteristic Curve):**

AUC-ROC evaluates the model’s ability to distinguish between classes across different decision thresholds by measuring the trade-off between the true positive rate (recall) and the false positive rate (FPR). Higher AUC values indicate better classification capability.

**vi. AUC-PRC (Area Under the Precision–Recall Curve):**

AUC-PRC measures the relationship between precision and recall across varying thresholds and is particularly useful for imbalanced datasets, where identifying minority classes correctly is critical.

## 4. Results and Discussion

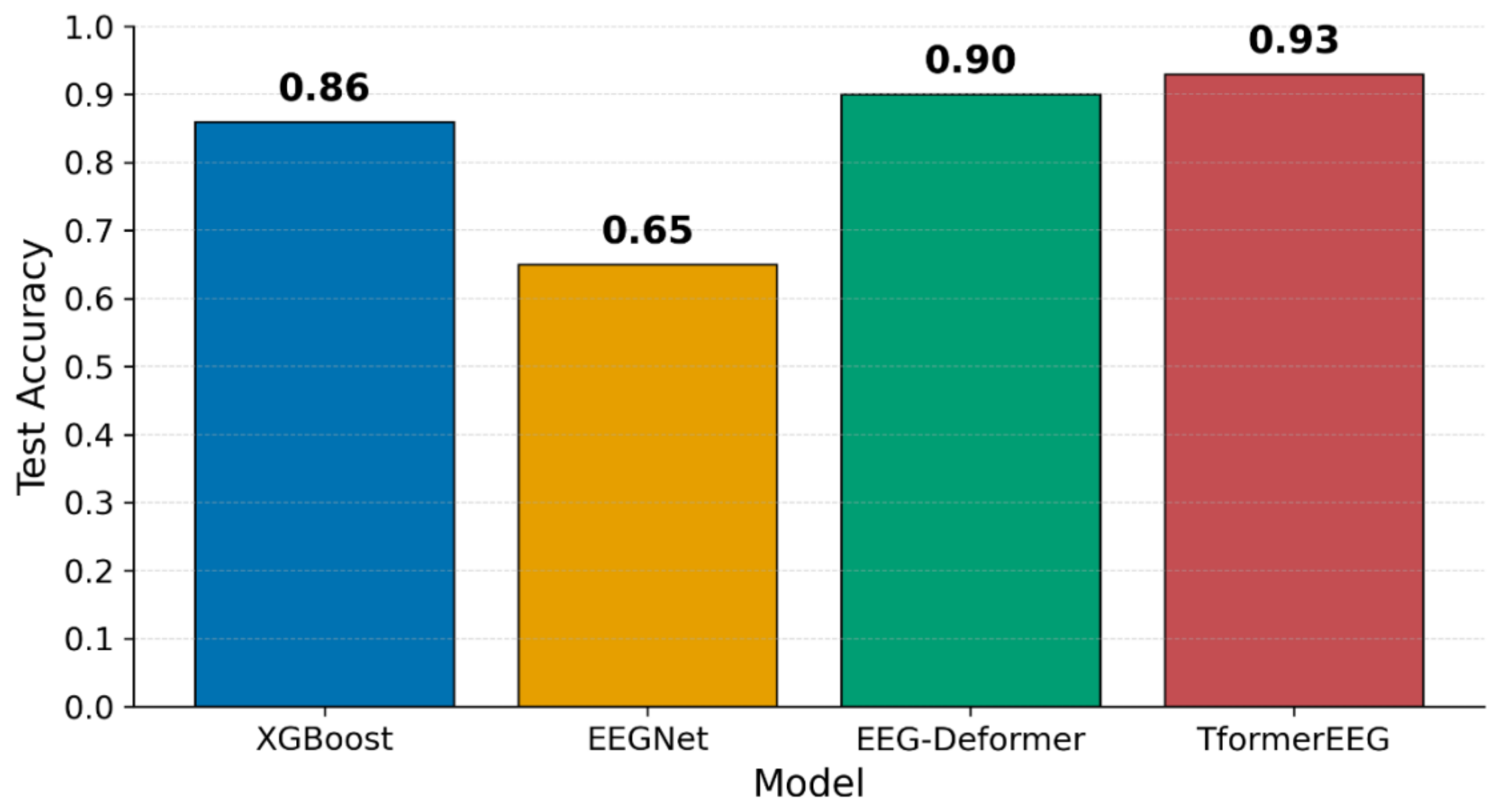

**Fig 7.** Comparison of Test Accuracy Across Models

The comparative analysis of test accuracy is shown in Figure 7. These demonstrate significant variation in their ability to accurately decode EEG motor imagery signals. Among them, the proposed TFormerEEG achieved the highest average test accuracy of 0.93, outperforming the EEG-Deformer model (0.90), XGBoost (0.86), and EEGNet (0.67). The detailed classification report of the models is presented in Table 3, where support denotes the number of test samples belonging to each class, Macro Avg represents the unweighted average of the evaluation metrics across all classes, and Weighted Avg represents the average weighted by the number of samples in each class. Notably, the TFormerEEG also maintained the lowest performance variability across folds (SD = 0.01), signifying greater stability and robustness in generalizing across different data partitions, as shown in the 10-fold cross-validation in Table 4. In contrast, EEGNet exhibited the lowest accuracy and the highest standard deviation (SD = 0.023), indicating inconsistent classification performance and limited generalization capacity.

From a performance dynamics perspective, the EEG-Deformer model proved effective, achieving a high average accuracy with minimal variability (SD = 0.0103), confirming the model's strength in capturing global dependencies via Multi-Head Self-Attention. XGBoost also performed reliably with consistent fold-wise accuracy, yet lacked the representational depth needed to extract complex spatiotemporal EEG patterns compared to neural sequence models.

The proposed TFormerEEG improves classification performance by effectively modeling the non-stationary temporal dynamics present in motor imagery EEG signals. Neural activity during motor imagery evolves sequentially over time, making the representation of temporal dependencies essential for accurate classification. TFormerEEG learns task-specific patterns directly from the EEG-derived input sequence. Its attention-driven design enables the model to capture long-range temporal dependencies and global contextual relationships within the EEG sequence, which is important for distinguishing subtle differences among rest, right-hand imagery, and left-hand imagery. This architecture enables the model to learn robust feature representations and achieve improved classification performance compared with conventional machine learning models such as XGBoost and baseline deep learning approaches, like EEGNet and the transformer-based model.

**Table 3.** Classification Report of the Model

| Model | Class | Precision | Recall | F1-Score | Support | AUC-ROC | AUC-PRC |
|---|---|---|---|---|---|---|---|
| **XGBoost** | 0 | 0.84 | 0.95 | 0.89 | 381 | 0.92 | 0.95 |
| | 1 | 0.89 | 0.79 | 0.84 | 193 | | |
| | 2 | 0.89 | 0.74 | 0.81 | 188 | | |
| | Accuracy | | | 0.86 | 762 | | |
| | Macro Avg | 0.87 | 0.83 | 0.84 | 762 | | |
| | Weighted Avg | 0.86 | 0.86 | 0.86 | 762 | | |
| **EEGNet** | 0 | 0.74 | 0.87 | 0.80 | 381 | 0.82 | 0.66 |
| | 1 | 0.57 | 0.51 | 0.54 | 193 | | |
| | 2 | 0.55 | 0.40 | 0.47 | 188 | | |
| | Accuracy | | | 0.67 | 762 | | |
| | Macro Avg | 0.62 | 0.60 | 0.60 | 762 | | |
| | Weighted Avg | 0.65 | 0.67 | 0.65 | 762 | | |
| EEG-Deformer | 0 | 0.88 | 0.94 | 0.91 | 381 | 0.972 | 0.957 |

| | 1 | 0.93 | 0.85 | 0.88 | 193 | | |
|---|---|---|---|---|---|---|---|
| | 2 | 0.91 | 0.88 | 0.89 | 188 | | |
| | Accuracy | | | 0.90 | 762 | | |
| | Macro Avg | 0.91 | 0.89 | 0.90 | 762 | | |
| | Weighted Avg | 0.90 | 0.90 | 0.90 | 762 | | |
| **TFormer EEG** | 0 | 0.94 | 0.95 | 0.95 | 381 | 0.982 | 0.967 |
| | 1 | 0.95 | 0.89 | 0.92 | 193 | | |
| | 2 | 0.89 | 0.93 | 0.91 | 188 | | |
| | Accuracy | | | **0.93** | 762 | | |
| | Macro Avg | 0.93 | 0.92 | 0.93 | 762 | | |
| | Weighted Avg | 0.93 | 0.93 | 0.93 | 762 | | |

**Table 4.** 10-Fold Cross-validation

| | **XGBoost** | **EEGNet [33]** | **EEG-Deformer [34]** | **TFormerEEG** |
|---|---|---|---|---|
| Fold 1 | 0.85 | 0.71 | 0.8924 | 0.9239 |

| Fold 2 | 0.86 | 0.67 | 0.8635 | 0.9265 |
|---|---|---|---|---|
| Fold 3 | 0.86 | 0.67 | 0.9003 | 0.9186 |
| Fold 4 | 0.82 | 0.67 | 0.8766 | 0.9160 |
| Fold 5 | 0.84 | 0.67 | 0.8766 | 0.9186 |
| Fold 6 | 0.85 | 0.73 | 0.8819 | 0.9055 |
| Fold 7 | 0.88 | 0.70 | 0.8766 | 0.9134 |
| Fold 8 | 0.82 | 0.69 | 0.8898 | 0.8924 |
| Fold 9 | 0.83 | 0.69 | 0.8921 | 0.9053 |
| Fold 10 | 0.85 | 0.74 | 0.8763 | 0.8974 |
| Mean | 0.85 | 0.69 | 0.8826 | 0.9118 |
| SD | 0.017 | 0.023 | 0.0103 | 0.010 |

For the statistical significance, paired t-tests were conducted. The statistical comparison results are shown in Table 5. The analysis indicates that the proposed TFormerEEG achieves statistically significant improvements compared with XGBoost ($t = 11.744$, $p < 0.001$), EEGNet ($t = 20.757$, $p < 0.001$), and the EEG-Deformer model ($t = 5.369$, $p < 0.001$). These results confirm that the proposed architecture consistently outperforms the baseline approaches under the evaluated experimental settings.

**Table 5.** Paired t-test results across 10-fold cross-validation

| Model Comparison | t-statistics | p-value | Significance |
|---|---|---|---|
| TFormerEEG vs XGBoost | 11.744 | 0.000000925 | $p < 0.001$ |
| TFormerEEG vs EEGNet | 20.757 | 6.54196964e-09 | $p < 0.001$ |
| TFormerEEG vs EEG-Deformer | 5.369 | 0.0004507 | $p < 0.001$ |

### 4.2 XGBoost

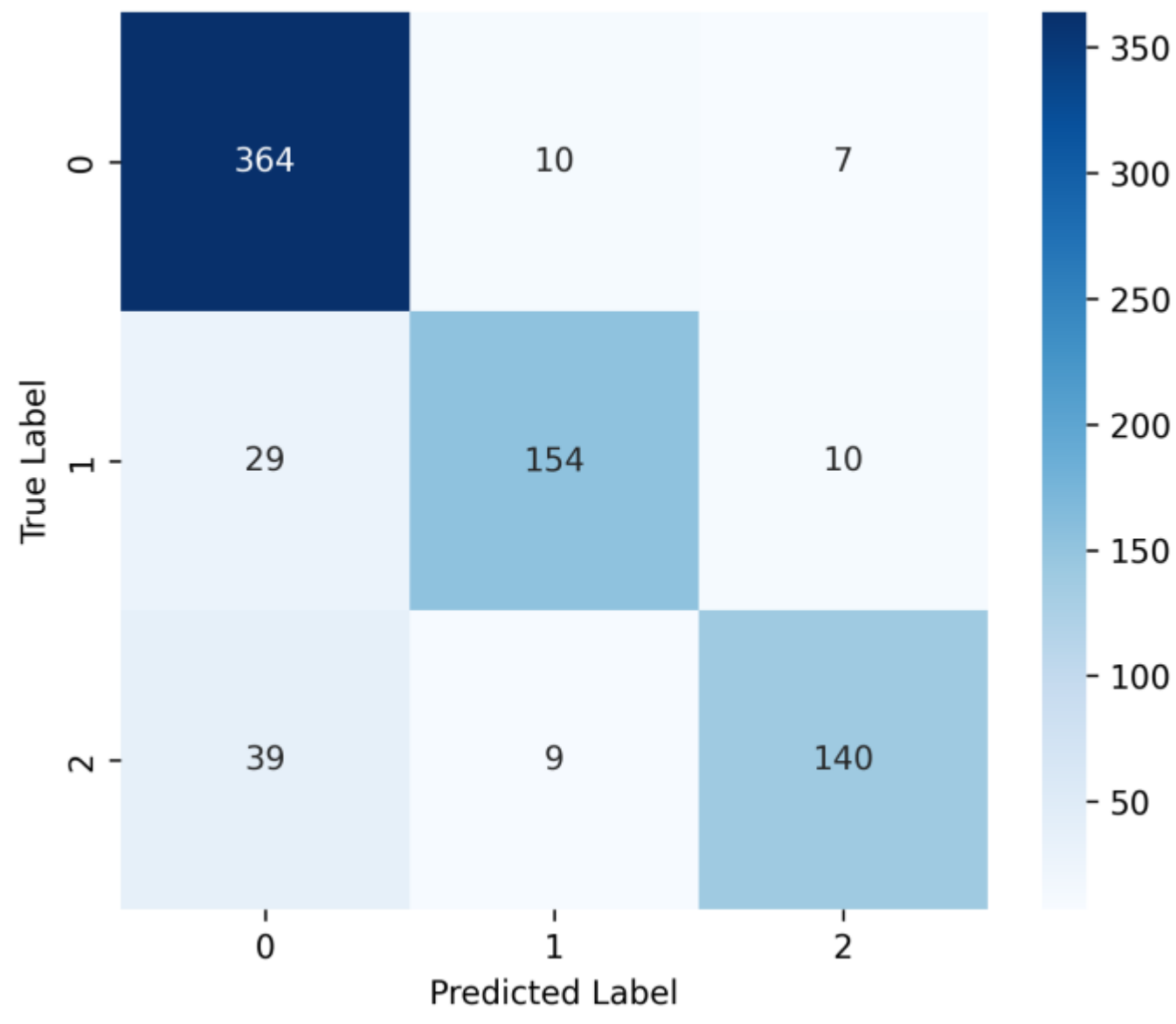

**Fig 8**. XGBoost Confusion Matrix

The XGBoost classifier, shown in Figure 8, achieved strong performance in classifying motor imagery EEG signals, with 364 of 381 instances correctly identified for class 0 (rest state), indicating effective separation from active motor tasks. For class 1 (right-hand imagery), the model correctly predicted 154 out of 193 instances, while for class 2 (left-hand imagery), 140 out of 188 were accurately classified. Most misclassifications occurred with class 0, suggesting potential overlap in EEG features between passive and active states, possibly due to transitional cognitive phases or inter-subject variability. Importantly, the model showed minimal confusion between left- and right-hand imagery, highlighting its ability to distinguish lateralized motor intent.

### 4.3 EEGNet

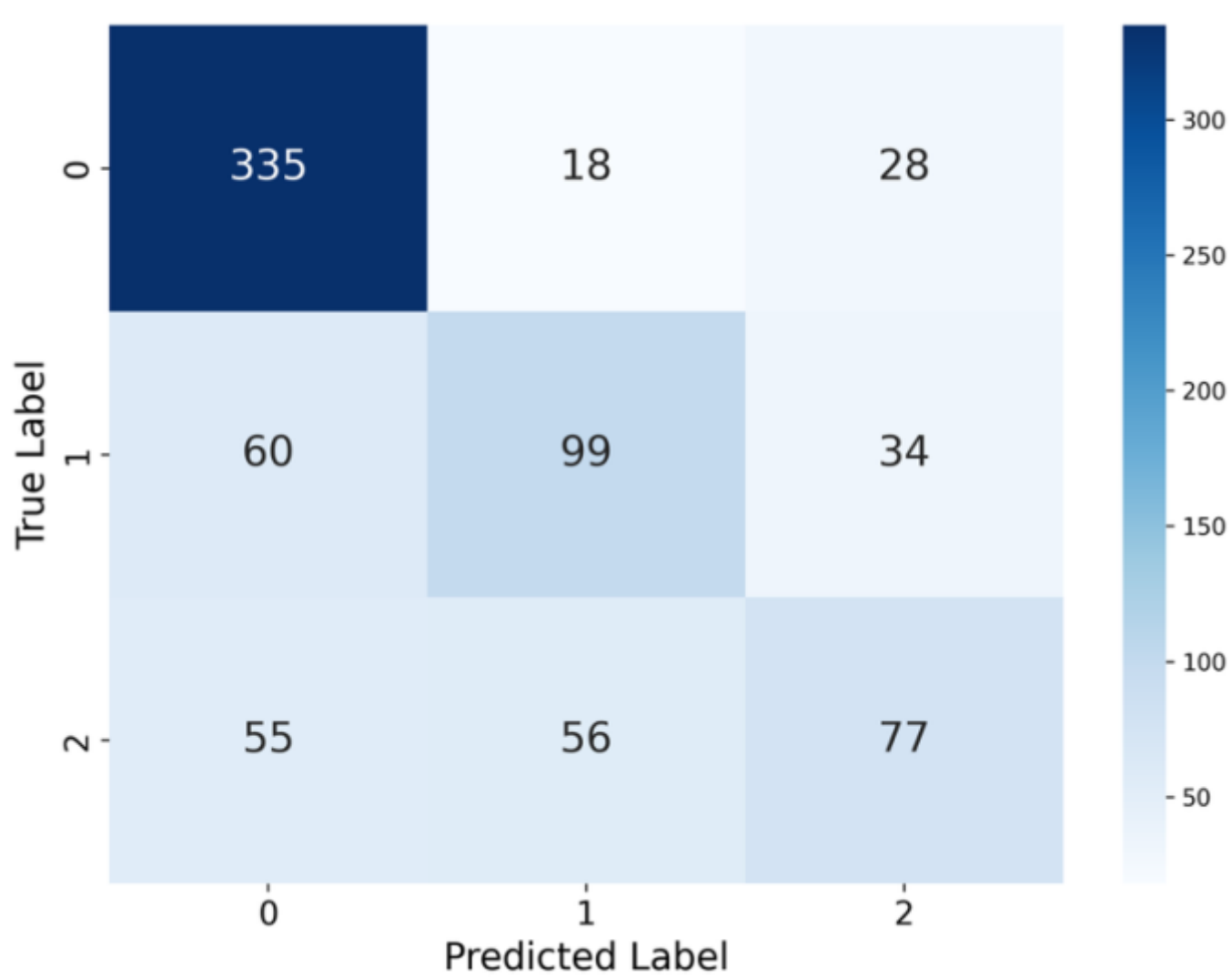

**Fig 9**. EEGNet Confusion Matrix

The EEGNet model exhibited moderate classification performance across the three motor imagery classes as shown in Figure 9. It achieved high accuracy for class 0 (rest), correctly predicting 335 out of 381 instances. However, classification performance declined for motor imagery tasks: class 1 (right-hand imagery) and class 2 (left-hand imagery) saw correct predictions of only 99 and 77 out of 193 and 188 samples, respectively. The high misclassification rates, particularly between motor imagery classes and rest, suggest that EEGNet struggled to fully capture the subtle spatiotemporal patterns inherent in imagined movements. This may be due to limited discriminative feature extraction in the shallower architecture of EEGNet, emphasizing the need for more expressive models or enhanced preprocessing pipelines in complex BCI tasks.

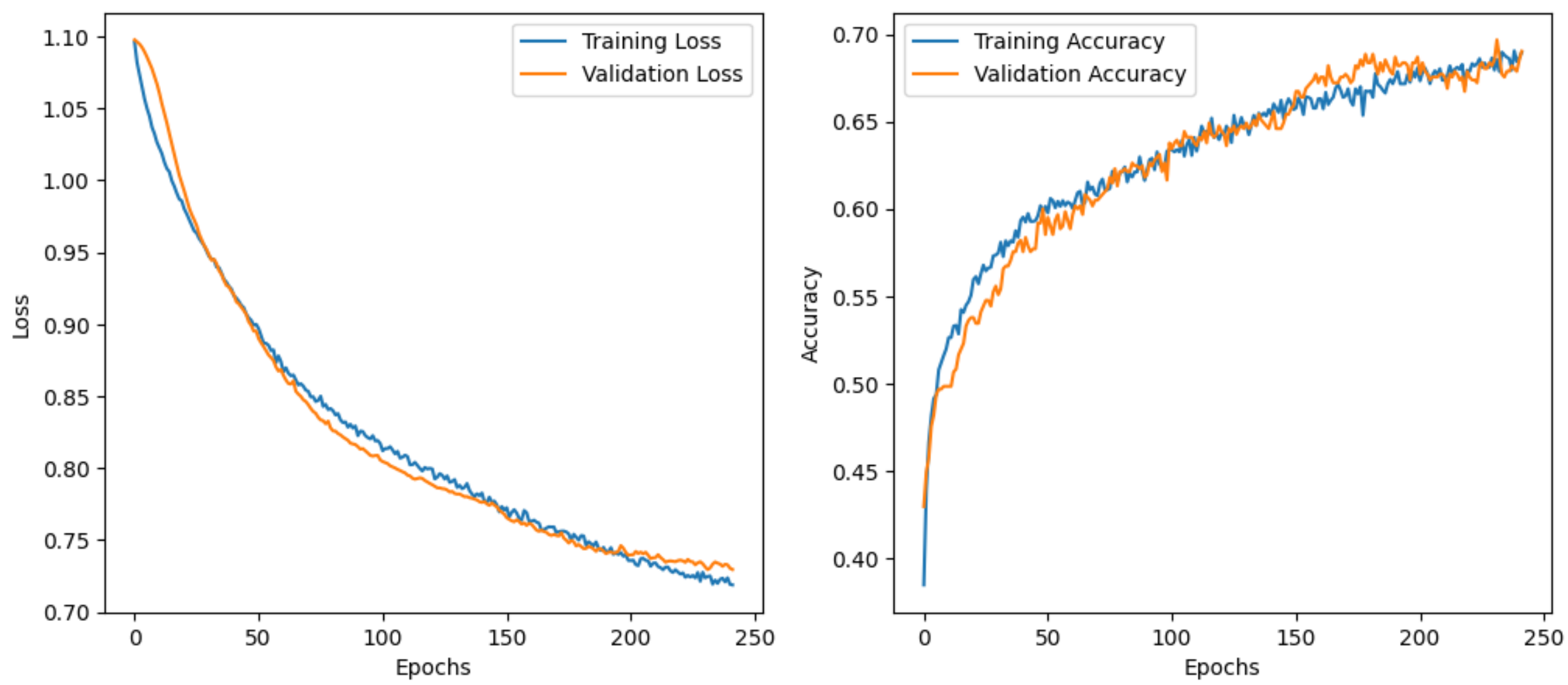

**Fig 10**. History Plot for EEGNet

The training history of EEGNet, as shown in Figure 10, indicates stable convergence over 250 epochs. Both training and validation losses exhibit a consistent decline, reflecting effective learning behavior. Concurrently, the accuracy curves demonstrate a gradual and steady increase, with validation accuracy closely aligning with training performance and attaining near 70%. This trend suggests a satisfactory level of generalization. Nevertheless, the observed performance plateau and comparatively modest accuracy indicate inherent limitations in the model's representational capacity to effectively capture complex EEG motor imagery signals, highlighting the need for more expressive model architectures or enhanced feature representation strategies.

### 4.4 EEG-Deformer

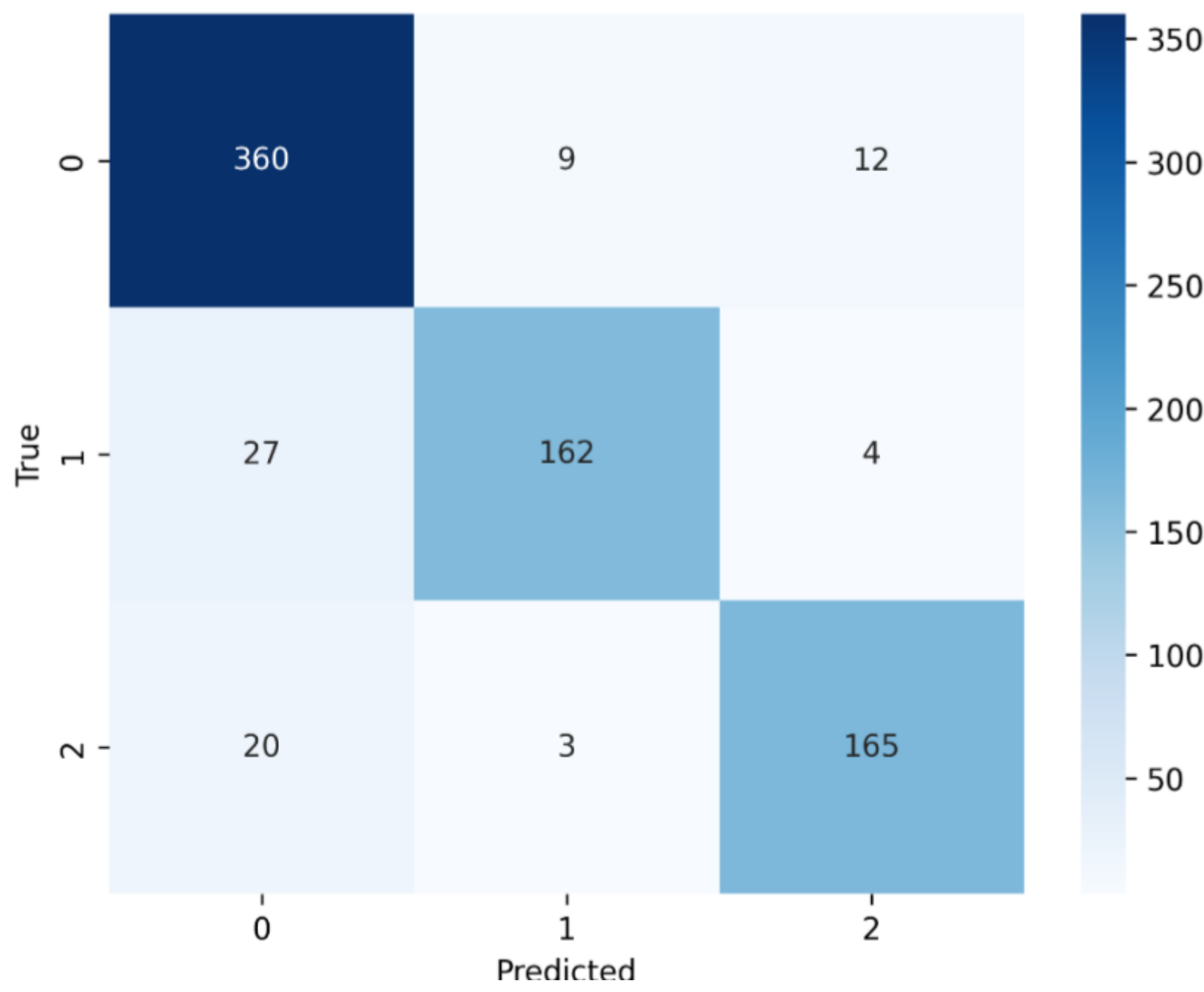

**Fig 11**. EEG-Deformer Confusion Matrix

The EEG-Deformer model demonstrates strong classification performance across all three EEG motor imagery classes, shown in Figure 11. Class 0 (rest) is predicted with high precision, with 360 correct out of 381 instances, and minimal misclassifications. Notably, the model achieves high accuracy for both motor imagery classes as well: class 1 (right-hand imagery) and class 2 (left-hand imagery) achieve 162 and 165 correct predictions, respectively. The low inter-class confusion—particularly between classes 1 and 2—indicates the model's effective capture of subtle spatiotemporal dependencies inherent in EEG signals.

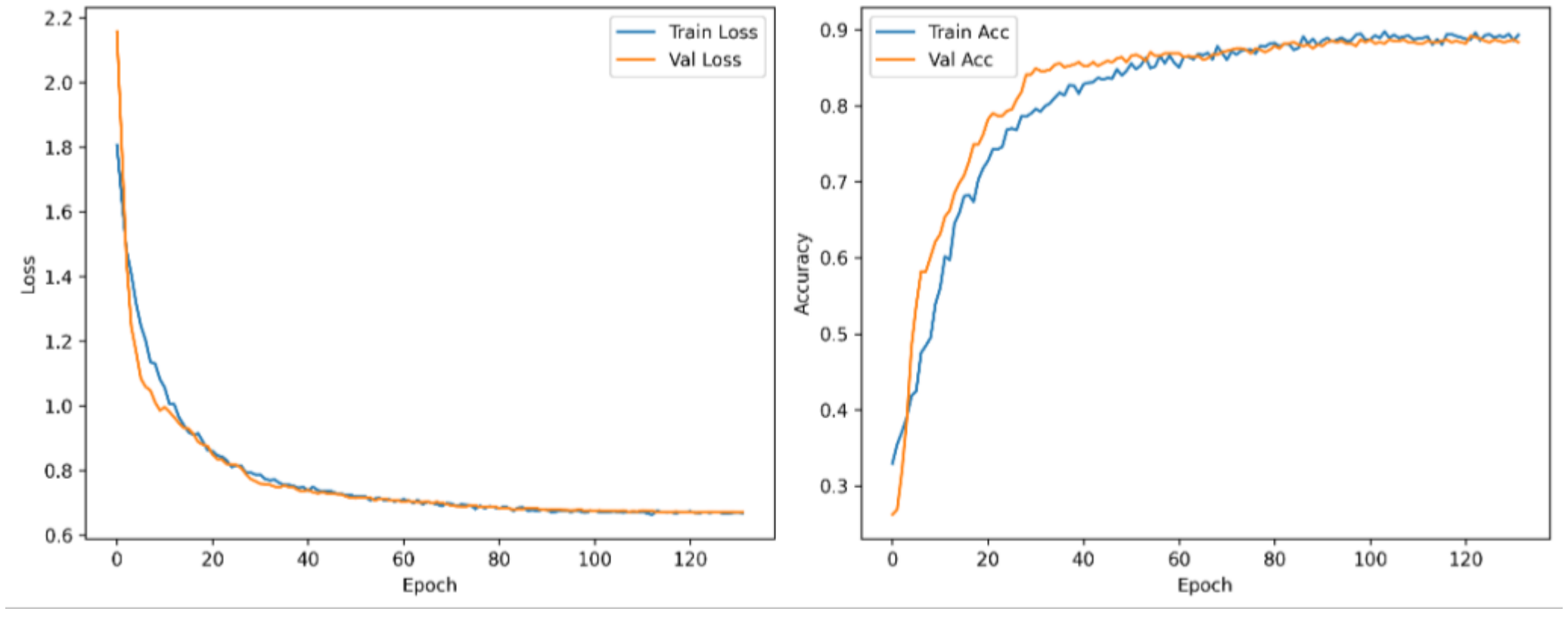

**Fig 12**. History Plot for EEG-Deformer

The training dynamics of the EEG-Deformer model, as shown in Figure 12, demonstrate effective learning and stable convergence. Both training and validation losses show a sharp decrease during the early epochs, followed by a gradual reduction until they converge at a similar low value near the end of training. The validation loss initially starts higher than the training loss but quickly follows the same downward trend, with no signs of divergence. The accuracy curves also show rapid initial improvement, where both training and validation accuracies increase steadily and then stabilize in the later epochs. The final training and validation accuracies remain closely aligned at approximately 89%, indicating good generalization with no clear evidence of overfitting. These results highlight the model's ability to learn discriminative EEG features effectively, supporting its robustness for EEG-based motor imagery classification tasks.

### 4.5 TFormerEEG

The proposed TFormerEEG model demonstrates superior performance in EEG-based motor imagery classification, as reflected in the confusion matrix in Figure 13. Class 0 (rest) is predicted with high precision, with 362 correct out of 381 instances, while class 1 and class 2 (right and left-hand imagery) achieve 172 and 175 correct predictions out of 193 and 188 samples, respectively. Misclassifications across classes are minimal, indicating effective spatiotemporal feature extraction. The model's ability to combine fixed-length EEG segmentation, token embedding, and self-attention enables effective learning of discriminative patterns and global contextual dependencies, which are essential for distinguishing subtle motor imagery patterns in EEG signals. This robust performance highlights the model's high discriminative capability and validates its suitability for real-time BCI applications requiring accurate, low-latency neural decoding.

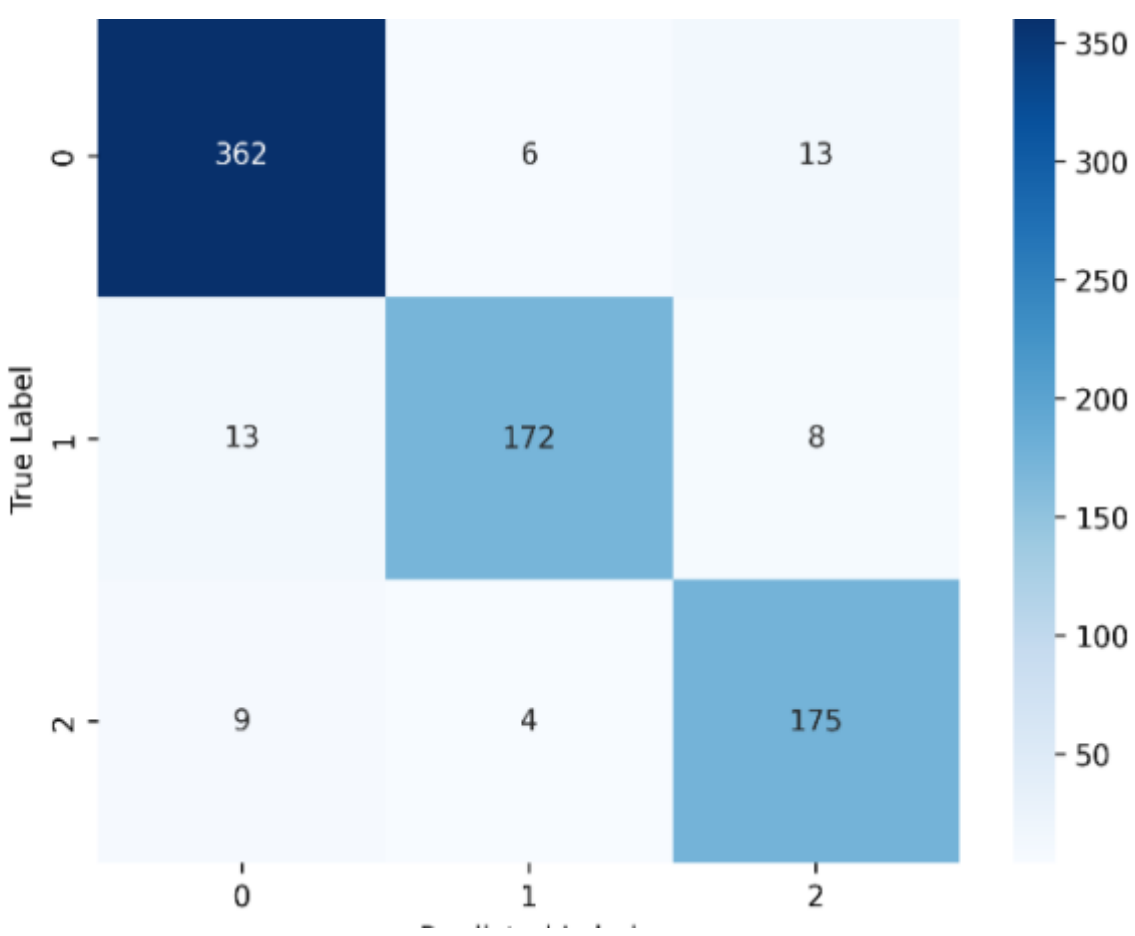

**Fig 13**. TFormerEEG Confusion Matrix

The training history of the proposed TFormerEEG reflects stable learning and good generalization capability, as shown in Figure 14. Both training and validation losses exhibit a smooth, steady convergence near zero by the end of training, indicating effective minimization of the objective function without instability. Training accuracy approaches approximately 95%, while validation accuracy stabilizes around 90–91% with minimal fluctuations, suggesting strong generalization. These results indicate that the TFormerEEG architecture effectively learns discriminative representations from EEG signals for motor imagery classification.

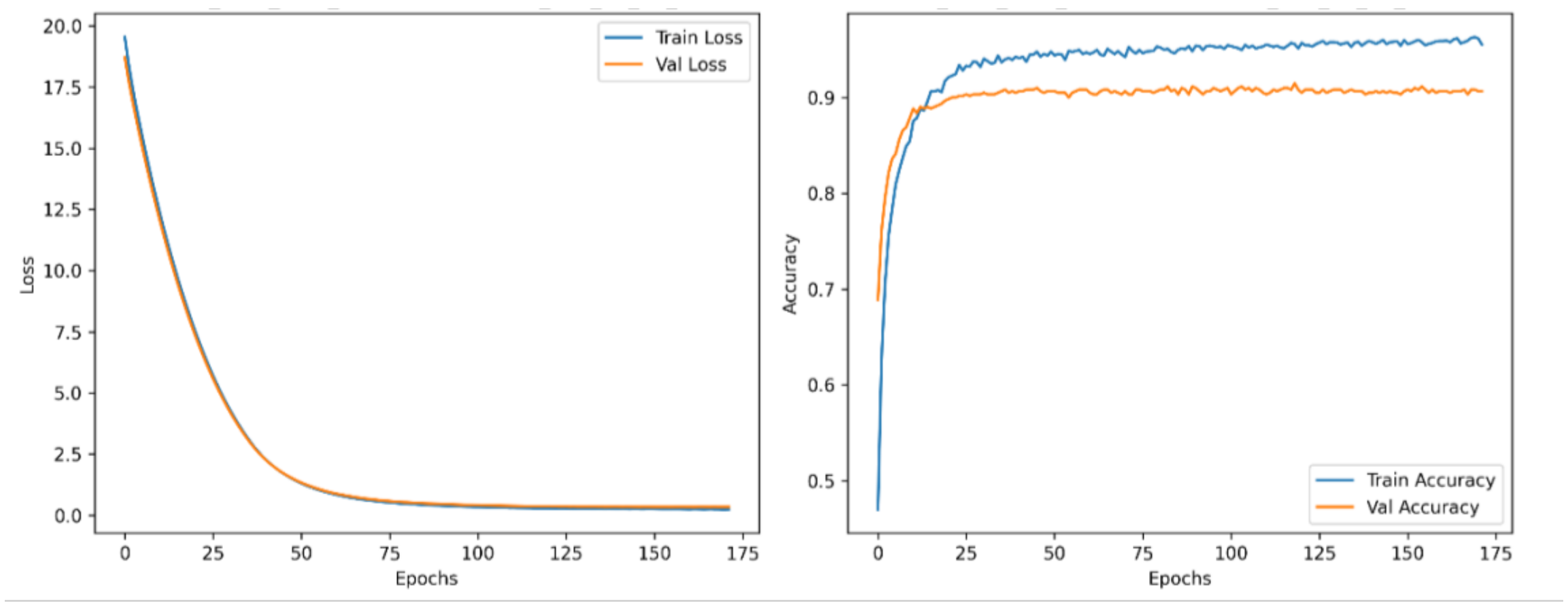

**Fig 14**. History Plot for TFormerEEG

## 4.6 Best hyperparameters for models:

The best hyperparameters for the model are presented in Table 6.

Table 6. Hyperparameters used for each model

| Model | Hyperparameters |
|---|---|
| **XGBoost Classifier** | n_estimators=300, learning_rate=0.01, objective='multi ', random_state=42, subsample=0.8, colsample_bytree=0.8, |
| **EEGNet** | F1=8, D=2, F2=16, kernLength=64, dropout rate=0.1, optimizer=Adam (learning rate=1e-4), batch size=128, epochs=250, early stopping (patience=10) |
| **EEG-Deformer** | Temporal Kernel=11, Depth=4, MaxNorm=1.0, Num Kernels=16, Heads=2, Dim Head=4, MLP Dim=4, Dropout=0.4, L2 Regularization=1e-4, Input Noise STD=0.01, Label Smoothing=0.2, Learning Rate=1e-4, Batch Size=32, Epochs=200, Early-Stopping Patience=10, ReduceLR Patience=4, ReduceLR Factor=0.5, Optimizer=Adam, Validation Split=0.2, Random Seed=42. |
| **TFormerEEG** | Patch size = 8, token embedding dimension = 16, learned positional encoding with embedding dimension = 16, multi-head attention = 4 heads with key_dim = 16, feed-forward network = 16 → 64 → 16, Gaussian noise = 0.01, dropout rate = 0.5, L2 regularization = 0.2 for patch embedding and classifier, L2 regularization = 0.3 for feed-forward layers, optimizer = Adam (learning_rate = 5e-5, clipnorm = 1.0), loss = sparse categorical cross-entropy, batch size = 32, epochs = 500, early stopping patience = 10, ReduceLROnPlateau (factor = 0.5, patience = 7, min_lr = 1e-5), validation split = 0.2. |

## 4.7 System simulation of wheelchair movement:

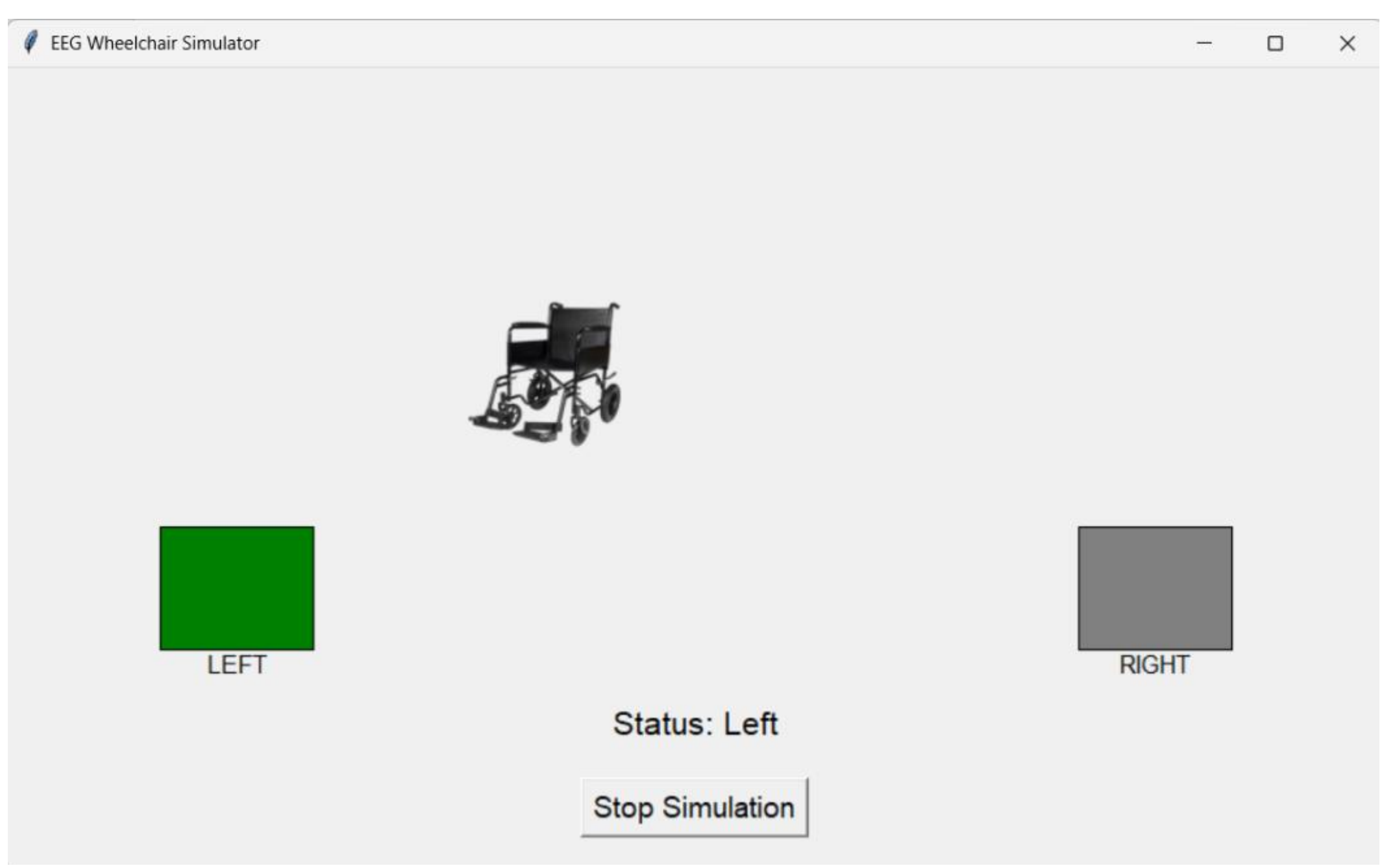

Fig 15 (a): System simulation in GUI (shows wheelchair left movement)

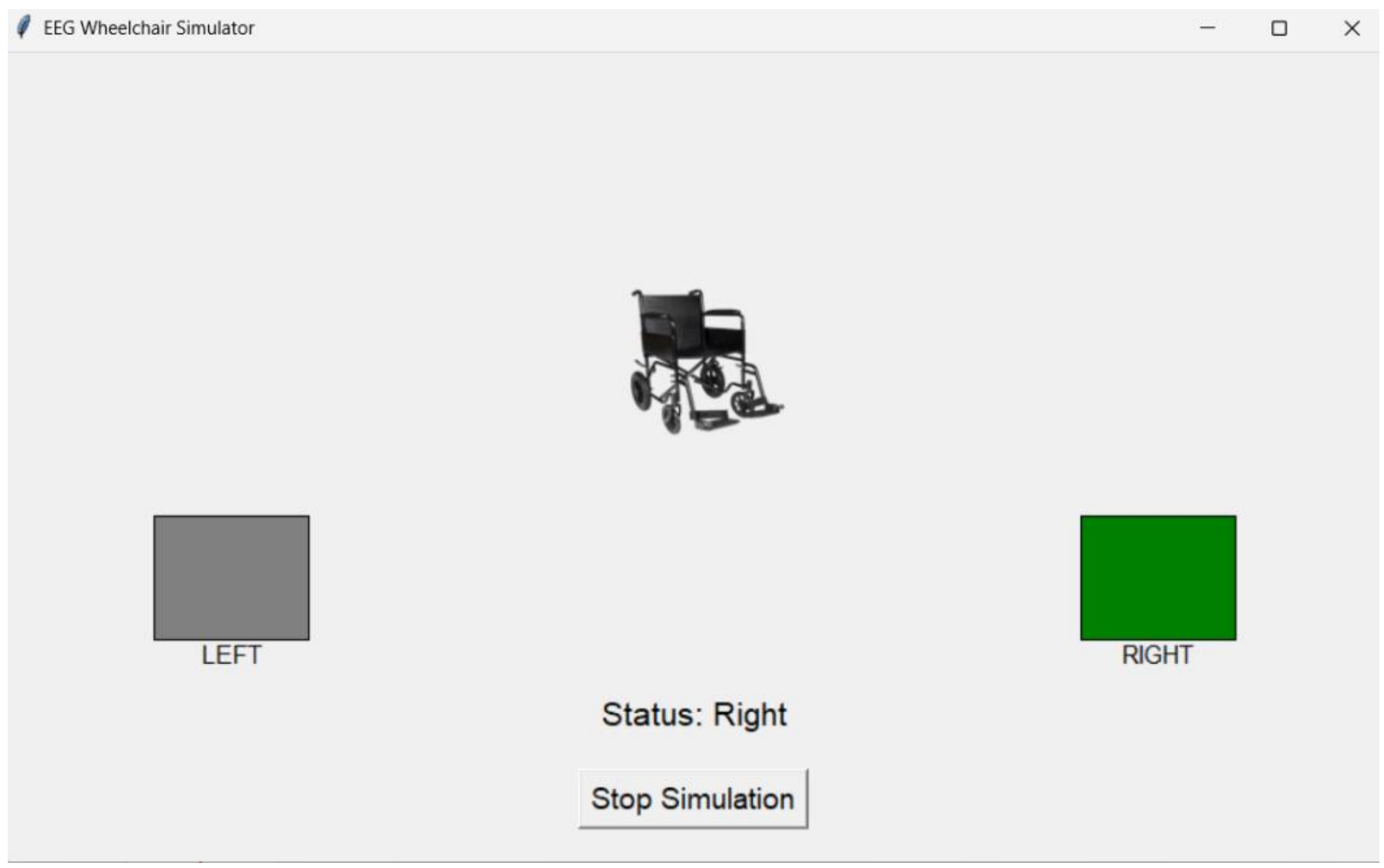

Fig 15 (b): System simulation in GUI (shows wheelchair right movement)

Figure 15 shows the final GUI for simulating the virtual wheelchair brain-computer interface. The backend of the tkinter is loaded with a TFormerEEG-trained .h5 extension model for simulating right or left direction movement. The wheelchair keeps moving as the model predicts '0' on test data fed with numpy, similarly, turns right when the model predicts '1', and finally, turns left when the model predicts '2'. This simulation shows how real-time wheelchair control can be made for a BCI model based on predicting the data from an EEG headset and controlling its movement. This model is implemented in the Raspberry Pi system as shown in the proposed circuit, which can be used to develop a BCI-based wheelchair. The deployed model on the Raspberry Pi showed a great result with the help of a testing set data array; the motors can be simulated to move right and left, showing potential for EEG right and left motor imagery, hand movement-powered motor control.

The system connection can be deployed in a real-time wheelchair system for the precise wheelchair movement using this technology, with an EEG dataset. The DC motor's movement using a motor driver indicates the possibility of successful deployment in a wheelchair-based system for real-time data acquisition, making it possible for the wheelchair system to move. Finally, the system simulates the Raspberry Pi microcontroller and motor driver-based wheelchair using the trained best model. The model integrated into Raspberry Pi in the form of the 'ONNX' version is used to simulate wheelchair movement on microcontroller devices.

### 4.8 Hardware Deployment Analysis

The trained TFormerEEG model is deployed on a Raspberry Pi to evaluate real-world performance. The hardware configuration is provided in Supplementary Section 1 Table S1, while the results of 20 inference runs that report the inference time, CPU usage, and RAM utilization are reported in Supplementary Table S2. The model achieved an average inference time of 0.0145 seconds, with stable resource usage (16.43% CPU and 257.4 MB RAM). These results validate that the proposed TFormerEEG can be deployed without significant computational overhead, making it practical for real-world systems.

### 4.9 Limitations and future scopes

Despite the promising results, this work is subject to implementation and methodological limitations. The experimental evaluation is conducted using data from a single subject (Subject E); therefore, the reported performance reflects subject-specific results and does not demonstrate cross-subject generalizability. Moreover, deep learning models inherently operate as complex black-box systems, which limit interpretability and pose reliability challenges in safety-critical biomedical applications. Additionally, the selection to deployment requires more hardware usage focus. Our work is compatible with Raspberry Pi, but comprehensive hardware integration for practical deployment remains incomplete. Moreover, despite successful real-time inference validation, a fully integrated powerful embedded computing platforms-based wheelchair system, including reliable stopping and safety mechanisms, is missing.

In future work, we aim to extend the proposed framework to multi-subject datasets and incorporate cross-subject validation to assess robustness for real-world assistive applications. We also plan to investigate explainable AI (XAI) techniques to improve the interpretability and reliability of the proposed model. We also aim to emphasize real-time motor control in a wheelchair system, including right and left-hand movement-based control, enhanced hardware integration, and safety validation for practical assistive deployment.

## 5. Conclusions

This paper presents a BCI-based wheelchair control system utilizing EEG motor imagery signals processed using TFormerEEG, a transformer-based deep learning model. The system achieved a mean 10-fold stratified cross-validation accuracy of 91.18% and held-out test accuracy of 93.04% within the evaluated subject. Comparative analysis against XGBoost, EEGNet, and EEG-Deformer models confirmed the superiority of the proposed approach. The work provides an effective foundation for advancing AI-integrated BCI systems for assistive mobility applications.

## Authors' Declarations:

***Funding:***

*The authors declare that no funding was received for this research.*

***Conflict of Interest:***

*The authors declare that they have no known competing financial interests or personal relationships that could have appeared to influence the work reported in this paper.*

***Ethics Approval Statement:***

*This research does not involve human participants or animals and utilizes already published datasets.*

***Data and Code Availability:***

*The original EEG recording used in this study is publicly available from Kaya et al. [32]( https://doi.org/10.1038/sdata.2018.211). The implementation of the proposed model and files are provided at the GitHub repository at: Biplov01/TFormerEEG.*

***Declaration of LLM's use:***

*The authors used LLM tools to minimize grammatical errors and enhance the quality of their writing.*

**Author Contributions**

Bipul Thapa (Conceptualization, Methodology, Software, Validation, Data curation, Formal analysis, Investigation, Visualization, Software, Writing—original draft, Writing—review & editing, Supervision, Project administration), Biplov Paneru (Methodology, Software, Validation, Formal analysis, Writing – original draft, Writing – review and editing, Investigation, Data curation), Bishwash Paneru (Visualization, Data Curation, Software, Formal analysis, Methodology, Investigation), Khem Narayan Poudyal (Methodology, Formal analysis, Validation, Investigation, Writing – review and editing, Supervision)

## Section 1: Hardware Configuration and Deployment Metrics

Table S1 presents the hardware configuration of the Raspberry Pi system used in the experimentation.

**Table S1: Hardware Configuration**

| Component | Available / Total |
| --- | --- |
| RAM | 1845.6 MB |
| Swap | 1845.0 MB |
| Disk | 14.03 GB |

Table S2 presents the results of 20 inference runs, reporting inference time, CPU usage, and RAM utilization during model deployment.

**Table S2: Inference performance metrics on Raspberry Pi (20 runs)**

| Row No. | Inference Time (s) | CPU Usage (%) | RAM Used (MB) |
|---|---|---|---|
| 1 | 0.0190 | 15.0 | 250.8 |
| 2 | 0.0176 | 15.0 | 257.6 |
| 3 | 0.0138 | 19.5 | 257.7 |
| 4 | 0.0138 | 17.1 | 257.8 |
| 5 | 0.0137 | 15.0 | 257.8 |
| 6 | 0.0140 | 15.4 | 257.8 |
| 7 | 0.0141 | 15.0 | 257.8 |
| 8 | 0.0138 | 17.5 | 257.8 |
| 9 | 0.0141 | 17.5 | 257.8 |
| 10 | 0.0143 | 19.0 | 257.8 |
| 11 | 0.0141 | 17.5 | 257.8 |
| 12 | 0.0140 | 15.4 | 257.8 |
| 13 | 0.0143 | 15.0 | 257.8 |
| 14 | 0.0140 | 15.4 | 257.8 |
| 15 | 0.0140 | 17.1 | 257.8 |
| 16 | 0.0144 | 17.5 | 257.8 |
| 17 | 0.0142 | 17.1 | 257.8 |

| 18 | 0.0144 | 17.1 | 257.8 |
|---|---|---|---|
| 19 | 0.0142 | 15.4 | 257.8 |
| 20 | 0.0140 | 15.0 | 257.8 |
| **Average** | **0.0145** | **16.43** | **257.4** |

### Section 2: Receiver Operating Characteristic and Precision–Recall Curves

This section presents the Receiver Operating Characteristic (ROC) and Precision–Recall (PRC) area under the curve for all evaluated models. The plots show one-versus-rest class-wise curves and the macro-averaged curve, providing a comprehensive evaluation of model performance across varying decision thresholds, particularly under class imbalance conditions.

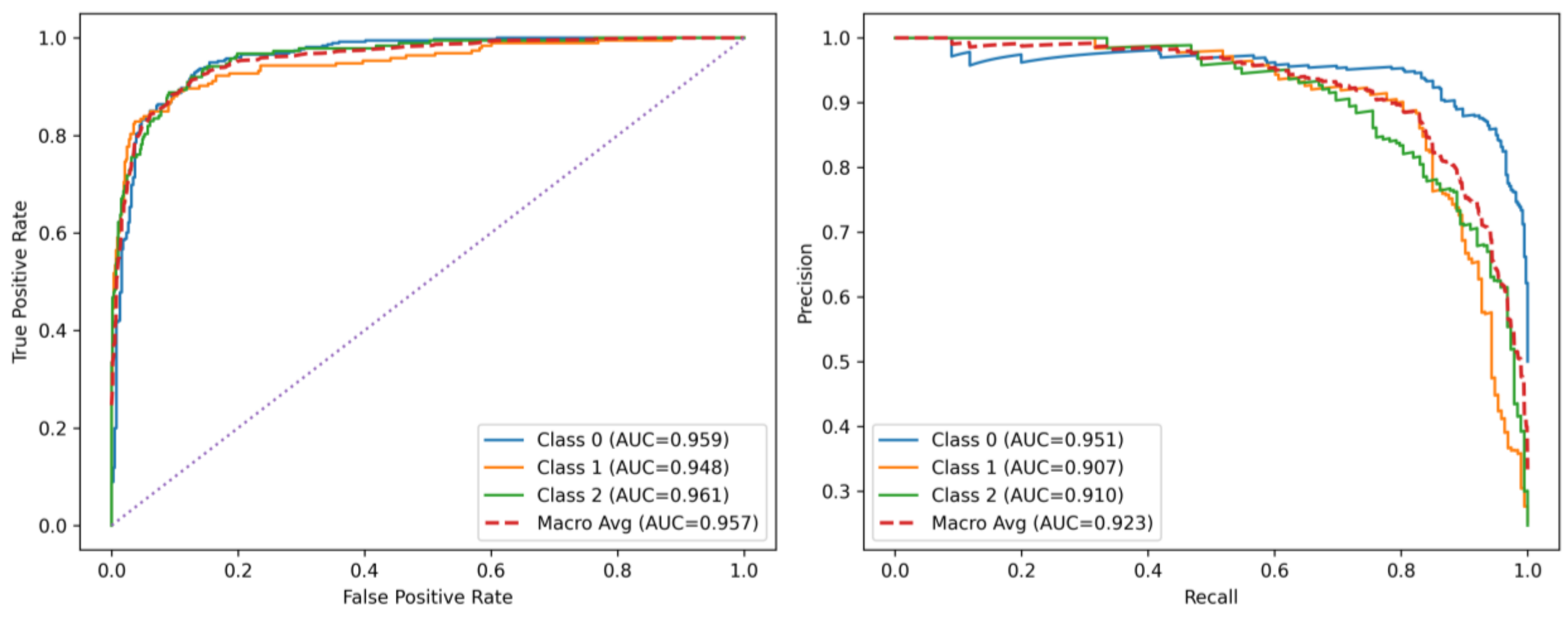

**Fig S1**. Performance curves for the XGBoost model: (a) ROC and (b) PRC

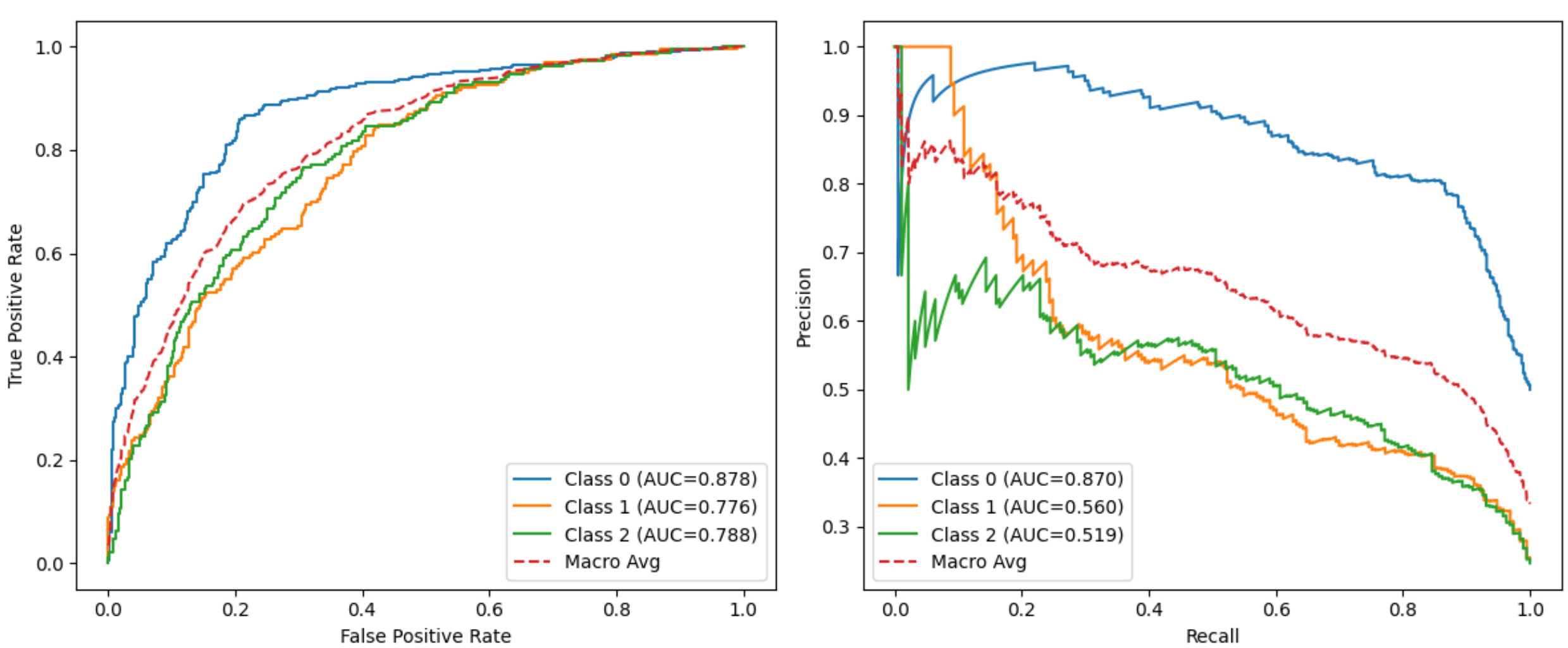

**Fig S2**. Performance curves for the EEGNet model: (a) ROC and (b) PRC

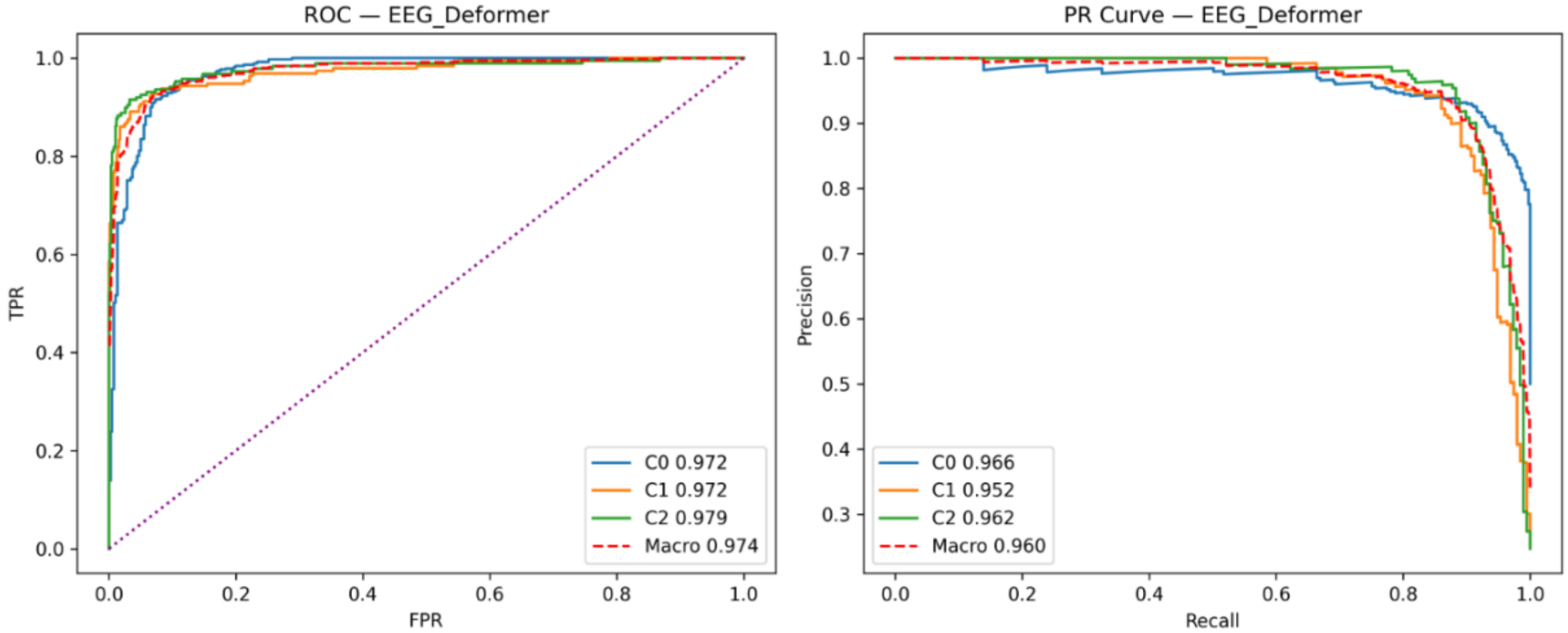

**Fig S3**. Performance curves for the EEG-Deformer model: (a) ROC and (b) PRC

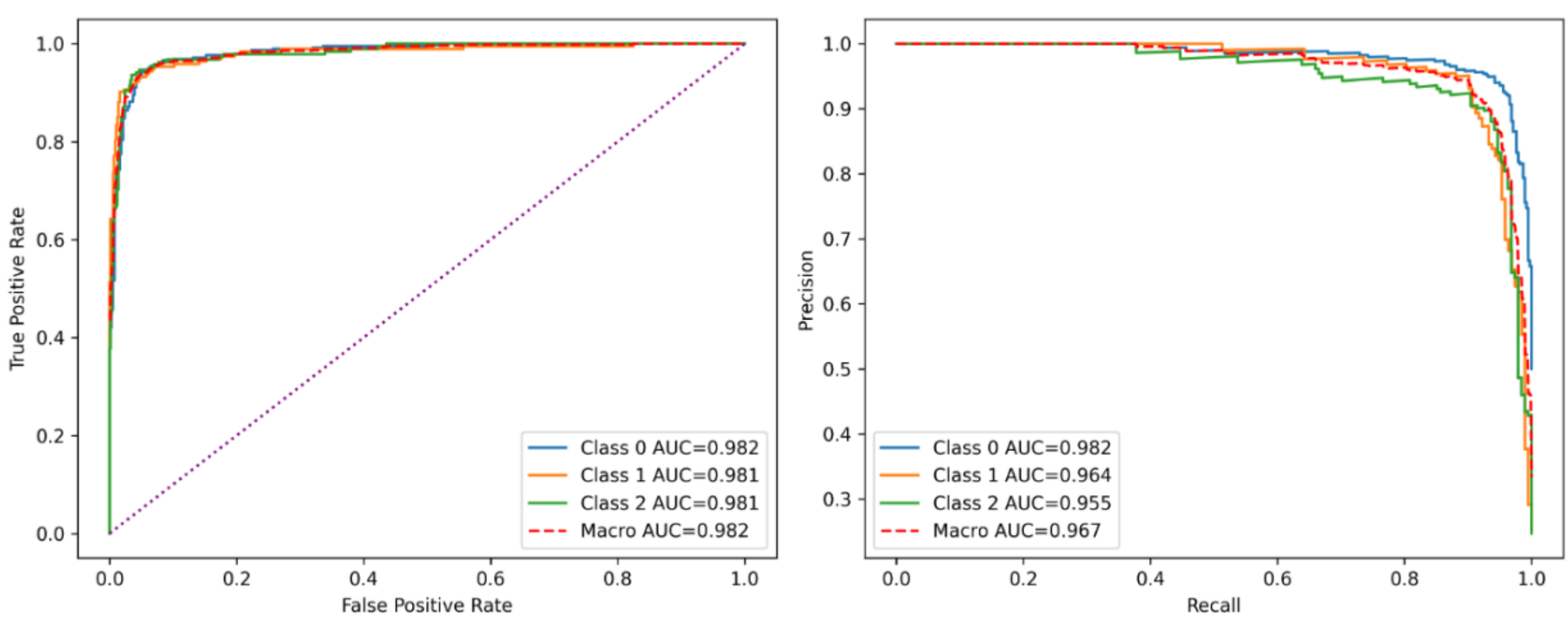

**Fig S4:** Performance curves for the TFormerEEG (a) ROC and (b) PRC